\documentclass{ck-report}
\definecolor{ckzebra}{HTML}{F5F3EE}    %
\definecolor{ckrowhair}{HTML}{D9D5CB}  %
\newcommand{\ckz}{\rowcolor{ckzebra}}                     %

\newcolumntype{Q}[1]{>{\raggedright\arraybackslash\rule{0pt}{10.5pt}}p{#1}<{\rule[-4.5pt]{0pt}{0pt}}}
\newcolumntype{W}[1]{>{\raggedright\arraybackslash\hsize=#1\hsize\rule{0pt}{10.5pt}}X<{\rule[-4.5pt]{0pt}{0pt}}}

\definecolor{ckrichallengefg}{HTML}{9A7B6C}\definecolor{ckrichallengebg}{HTML}{F3ECE6}
\definecolor{ckriinterventionfg}{HTML}{6F86A0}\definecolor{ckriinterventionbg}{HTML}{E9EEF4}
\definecolor{ckrimeasuredfg}{HTML}{5F8D86}\definecolor{ckrimeasuredbg}{HTML}{E6F1EF}
\definecolor{ckrilessonfg}{HTML}{A9502F}\definecolor{ckrilessonbg}{HTML}{F6E6DD}
\definecolor{cklessonbg}{HTML}{FCF6F1}
\newtcbox{\ckchipbox}[1][]{on line, box align=base, boxrule=0pt, arc=1.5pt, boxsep=0pt,
  left=2.2pt, right=2.2pt, top=1pt, bottom=0.8pt, #1}
\newcommand{\ckri}[2]{\ifstrequal{#1}{lesson}{\cklessonchip}{{\color{ckri#1fg}\sffamily\bfseries\fontsize{6.8}{8}\selectfont\textls[80]{\MakeUppercase{#2}}}\,}}
\newcommand{\cklessonchip}{\ckchipbox[colback=ckrilessonbg, coltext=ckrilessonfg]{%
  {\sffamily\bfseries\fontsize{6.6}{8}\selectfont\textls[80]{LESSON}}}\hspace{0.35em}}
\newtcolorbox{cklesson}{enhanced, breakable=false, colback=cklessonbg, colframe=ckclay!30!white,
  boxrule=0.4pt, arc=2pt, left=6pt, right=6pt, top=3pt, bottom=3pt, boxsep=0pt,
  before skip=4pt, after skip=4pt, fontupper=\small}

\newcommand{\freezedate}{\textcolor{red}{[FREEZE-DATE]}}
\newcommand{\boarddate}{\textcolor{red}{[BOARD-DATE]}}

\renewcommand{\boarddate}{2026-11-11}

\newcommand{\quantrank}{\#1}
\renewcommand{\freezedate}{2026-11-11}

\newcommand{\demourl}{https://huggingface.co/spaces/QianQu/CompoundingKernel-Explorer}

\title{CompoundingKernel: Self-Improving from Zero to First}
\hypersetup{pdftitle={CompoundingKernel: Self-Improving from Zero to First}, pdfauthor={CompoundingKernel team}}
\author{Qian Qu, Hongyang Wang, Wenbo Chen, Yichuan Li, Jingguo Liu, Xian Li, Xuan Wang,\\
  Ruochen Wen, Leo Fan, Yiming Yan, Yutong Chen, Zheshen Wang, Benoit Dumoulin}
\ckorg{Amazon}
\cklinks{{\color{ckink2}\faCode}\hspace{0.2em}\href{\demourl}{HuggingFace Space}}
\ckcorrespondence{Correspondence: Qian Qu \texttt{<quqian@amazon.com>}, \texttt{<eequq99@gmail.com>}.}

\begin{document}
\maketitle

\begin{abstract}
We present CompoundingKernel, a self-improving agent system that writes GPU kernels by search. As of \boarddate{}, its generated kernels rank \#1 on the public L1, L2 and Quantization boards of SOL-ExecBench (NVIDIA B200). They have held \#1 on L1 since 2026-10-27, albeit a short loss on 2026-11-07. All 235 problems have a site-verified correct kernel; on the 209 public-board problems they are 4.40$\times$ faster than the benchmark's reference baseline. The search starts from zero: agents get each problem's PyTorch definition but no reference kernels or official scores. A per-problem archive outside the model holds every correct kernel and its lineage. An LLM controller reads it, picks a parent or a new design, and gives each parallel child agent one direction. Checks outside the model decide correctness. At equal wall time, it beat a coded UCB style tree on 35 of 38 problems by making 2.4$\times$ as many correct attempts. Over a thousand agents ran at once, and one warm restart fix raised their evaluation rate 3.6$\times$. We find that the rate of reliable measurements, not the number of agents, limits how far such a search scales. Our code and kernels are available on \href{\demourl}{URL}.
\end{abstract}

\vspace{-0.6\baselineskip}\par\noindent\begin{minipage}{\linewidth}
  \centering
  \includegraphics[width=0.85\linewidth]{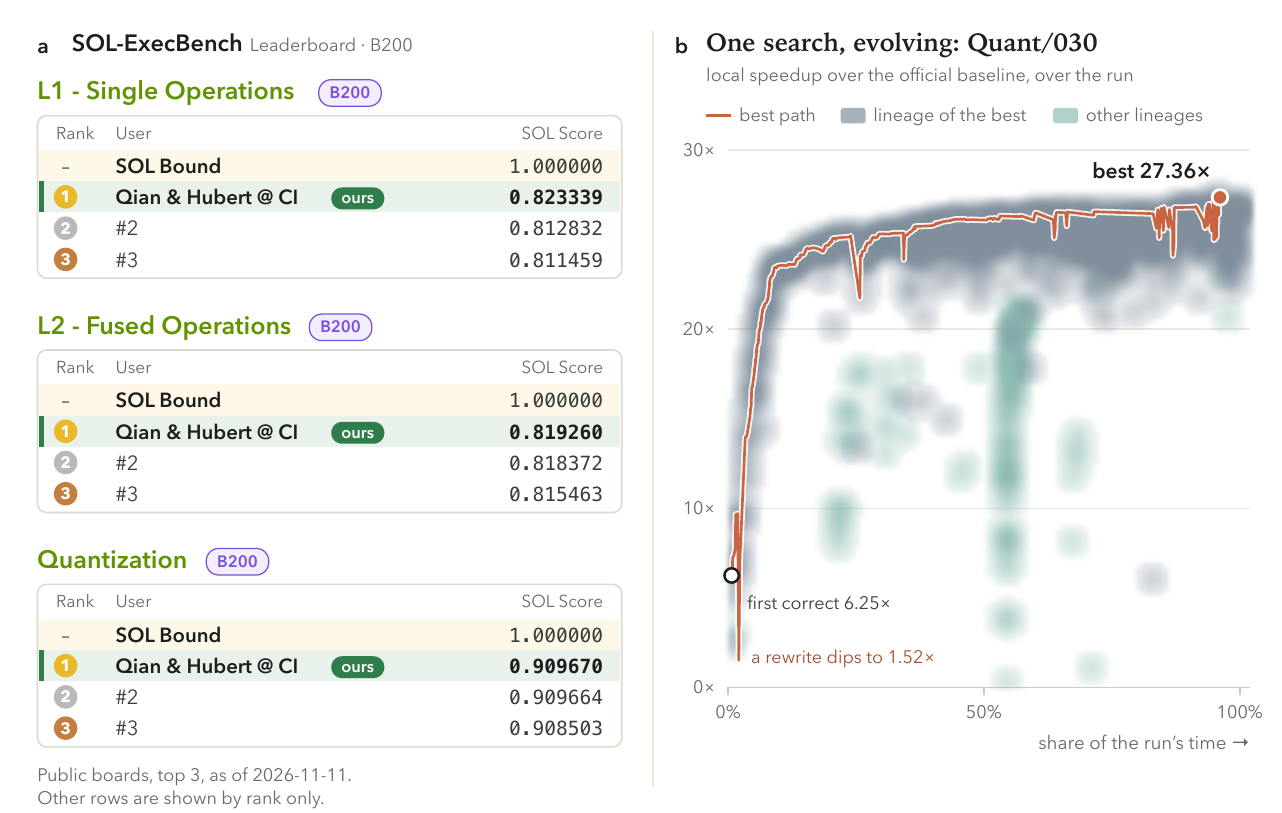}
  \captionof{figure}{We rank \#1 on L1, L2 and Quantization, through searches that compound. (a) Public boards, top 3, as of \boarddate{}; (b) Quantization board problem No.030's best path, 6.25$\times$ to 27.36$\times$, over its lineage (grey) and others (teal).}
  \label{fig:F1}
\end{minipage}\par\vspace{0.2\baselineskip}

\section{Introduction: From Zero to \#1}
\label{sec:intro}

A kernel is the program that runs one operation on a GPU, and faster kernels make the same model work take less GPU time. New hardware features arrive faster than libraries can use them. On NVIDIA Blackwell, the fast paths go through the \texttt{tcgen05} tensor-core instructions, tensor memory and the Tensor Memory Accelerator (TMA) copy engine, and most fused or low-precision shapes have no library kernel that uses them yet. Writing such a kernel is a search: try a design, measure it, fix what the measurement shows, and try again. The hard part is making that search build on its earlier tries, run it across hundreds of problems at once, and keep it from fooling itself.

CompoundingKernel is our system for this search. It starts each problem from its PyTorch definition, with no reference kernels, and official scores never enter the search. As of \boarddate{}, its kernels rank \#1 on the three public SOL-ExecBench boards we entered: L1 (0.823339), L2 (0.819260) and Quantization (0.909670; Figure~\ref{fig:F1}). These boards rank teams by their mean score over a whole collection of problems. The site also ranks each problem on its own, and there we are first on 89 of the 209 problems we entered and in the top 10 on 193. Every one of the 235 problems has a kernel that the site marks correct\footnote{FlashInfer-Bench kernels were verified correct on the site as private rows; we do not publish their results.}. On the 209 problems of the three public boards, these kernels are 4.40$\times$ faster than the benchmark's PyTorch baseline (geometric mean; median 3.78$\times$; site timing of our public rows, as of \boarddate{}). Over a thousand agent sessions ran at once, and 98.4\% of the candidates they evaluated were correct (as of 2026-11-02).

``From zero'' describes the search, not the model. Each run starts cold, with the problem's PyTorch definition, shapes and data types and nothing from earlier runs, and it learns only from our own tests and timings. It never sees a reference kernel or an official result. The model itself was pretrained on public code, and agents could read the CUTLASS headers, the web and a public wiki (\S\ref{sec:cleanroom}).
Figure~\ref{fig:F1}b shows one such search, on Quant/030, climbing from 6.25$\times$ to 27.36$\times$ (local timing). The \emph{harness} around the agents (their prompts, tools and evaluation) changed only at \emph{warm restarts}. A warm restart stops the agents, updates the harness and relaunches them, but keeps the run and every problem's archive of past attempts. The agents keep no memory of their own, so after the restart they re-read the archive and the search continues where it stopped, rather than starting from zero as in a cold start.

\begin{figure}[!htbp]
  \centering
  \includegraphics[width=\linewidth,height=0.45\textheight,keepaspectratio]{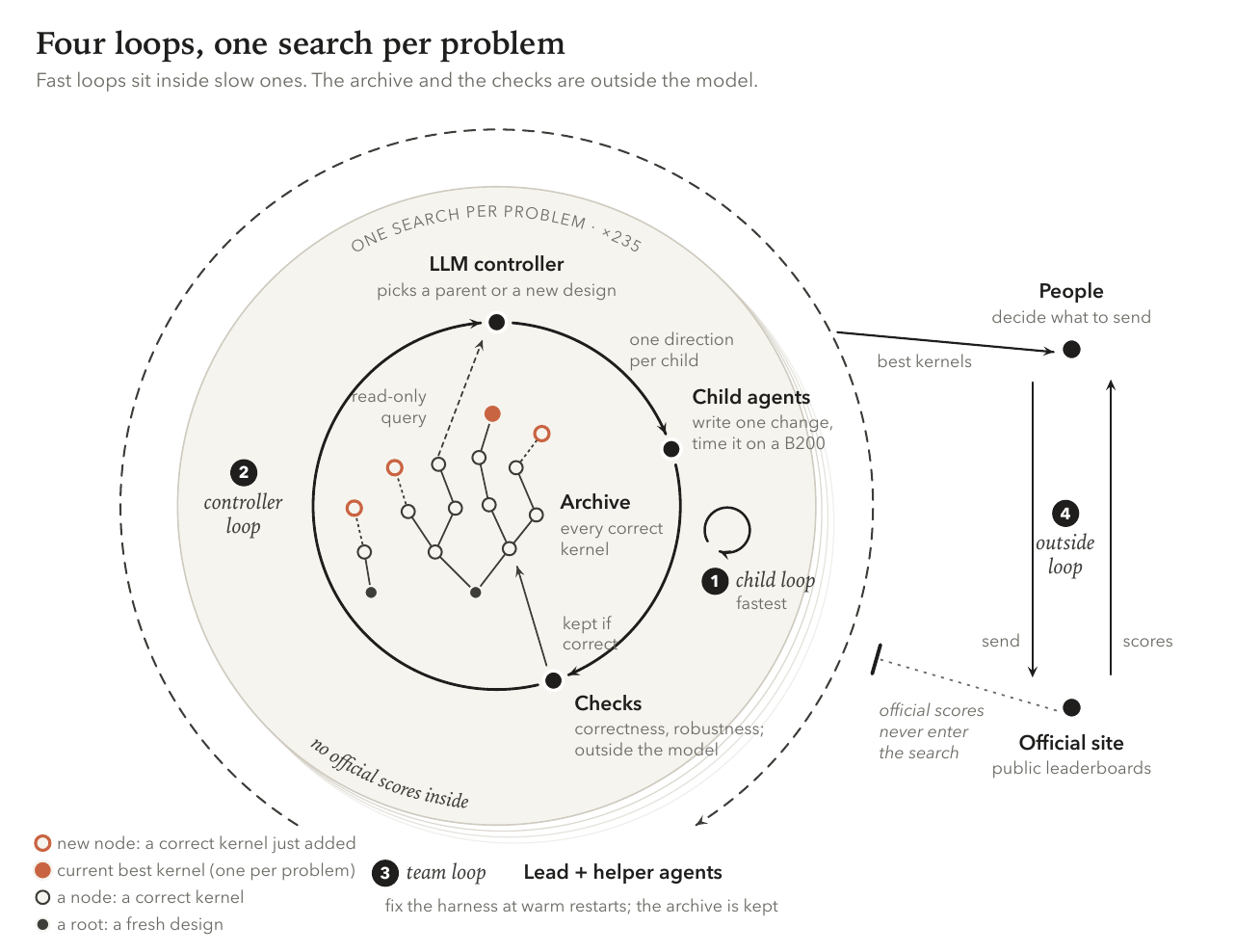}
  \caption{Four loops, one search per problem: \circled{1} a child writes one change, \circled{2} the LLM controller picks what next, \circled{3} agents fix the harness at warm restarts, \circled{4} people decide sends. Official scores never enter the search; people use the standings outside it, to choose sends and to place compute.}
  \label{fig:F2}
\end{figure}

The system is built as four nested loops (Figure~\ref{fig:F2}), each slower than the one inside it. In the innermost loop \circled{1}, a \emph{child} agent is given a starting kernel (its \emph{parent}) and one direction to try; it edits, compiles and times the code on a B200, checks outside the model decide whether the result is correct, and the new candidate is written with its timings and parent to the problem's \emph{archive}, a record of every kernel tried so far. In the controller loop \circled{2}, whenever a child slot frees up, an LLM controller reads the archive, picks an existing kernel to refine or opens a new design, and writes the direction the next child will follow; because the archive keeps every correct kernel, a slow kernel can later become the parent of the fastest one (\S\ref{sec:cases}). In the team loop \circled{3}, lead and helper agents monitor the system and progress, fix faults in the harness, notify the people for decisions if needed and each fix goes live at the next warm restart without losing the archives. In the outer loop \circled{4}, people decide which kernels to send to the public board and when. Official scores never enter loops \circled{1} and \circled{2}, which learn only from our own tests and timings.

Most strong recent kernel results on new hardware come from a fixed frontier model inside a search or harness~\citep{chen2026avo,chen2026forgekernel,nvlabs2026kda,cursor2026multiagentkernels}. A goal-only agent is already a strong baseline~\citep{chen2026justgoal}, so a harness has to earn its place by measurement. Ours did so in two ways. At equal wall time, our LLM controller beat a coded tree search that ranks nodes with an upper-confidence-bound (UCB) style score on 35 of 38 paired problems. Per attempt the two picked about equally well; the controller won because it made 2.4$\times$ as many correct attempts (\S\ref{sec:lesson-5}). Across the whole fleet, one warm restart fix raised evaluated candidates per hour 3.6$\times$ (\S\ref{sec:lesson-1}). More guidance, by contrast, did not help: making our optional references (our skill and a public wiki) mandatory added nothing measurable (\S\ref{sec:design-guidance}). Extra computation tends to win~\citep{sutton2019bitter}, but in kernel search it pays off only when each new try builds on a reliable measurement of the last one. The rate of such measurements, not the number of agents, is what limited scale.

The rest of the report follows this story. Section~\ref{sec:challenge} introduces the benchmark and its score. Section~\ref{sec:approach} describes CompoundingKernel: an archive outside the model, LLM node selection, parallel children that each follow one direction, and checks outside the model that decide correctness and caught reward hacks during the run (\S\ref{sec:verify}). Section~\ref{sec:results} gives the board standings and speedups, each with its timing basis and date. Section~\ref{sec:lessons} draws four lessons from running the system at scale, and Section~\ref{sec:design} gives three findings on search design, two of them null results. Section~\ref{sec:cases} follows searches lineage in which a slower kernel became the base of the best one. We plan to release our code (search loop, SOL-ExecBench interface, checks, example configuration) and our kernels (\S\ref{sec:conclusion}).

\nobarriersection
\section{The Challenge: SOL-ExecBench on Blackwell}
\label{sec:challenge}
\restorebarrier

SOL-ExecBench~\citep{lin2026solexecbench} is NVIDIA's public benchmark for kernels written by AI agents. Its 235 problems come from 124 real models; on the official boards they cover 3,957 workloads (input shapes). The workloads include forward and backward passes in BF16, FP8 and NVFP4. The problems fall into four collections, each with its own board: L1 (single operations, 94 problems), L2 (fused operations, 82), Quantization (33) and FlashInfer-Bench (26)~\citep{xing2026flashinferbench}.

\paragraph{The score.} Each workload $w$ in a problem's set of workloads $\mathcal{W}$ has two published times. The first is $t_b(w)$, the time of an optimized PyTorch baseline that NVIDIA has not released. The second is $t_\mathrm{sol}(w)$, the speed-of-light bound: the time a perfect kernel would take on a B200~\citep{nvlabs2026solar} running at the hardware limit. Let $t_k(w)$ be the measured time of kernel $k$ on workload $w$. The SOL score of $k$ on $w$ is

\[
  S_w(k) = \frac{t_b(w) - t_\mathrm{sol}(w)}{\bigl(t_k(w) - t_\mathrm{sol}(w)\bigr) + \bigl(t_b(w) - t_\mathrm{sol}(w)\bigr)}.
\]

A kernel that matches the baseline scores 0.5, a kernel at the bound scores 1, and a failed workload scores 0. A problem's score is the arithmetic mean over its workloads, and a board's score is the mean over its problems. The site checks correctness on every workload and then times the kernel on its own B200, with the streaming-multiprocessor (SM) clock locked at 1,500 MHz. It stops any build that takes longer than 300 s, and it refuses solutions that build native code at run time. We use four metrics (Table~\ref{tab:metrics}) and always label which one we mean.

\begin{table}[!htbp]
  \centering\cktablesetup\renewcommand{\arraystretch}{1}
  \caption{Metrics in this report. Local timings run at a higher clock than the site's and read fast (\S\ref{sec:reward}).}
  \label{tab:metrics}
  \begin{tabularx}{\linewidth}{@{\hspace{4pt}}W{1}W{1}W{1}@{\hspace{4pt}}}
    \toprule
    \thead{Metric} & \thead{Timed where} & \thead{Used for} \\
    \midrule
    Site SOL score & official site & board ranks (\S\ref{sec:boards}) \\
    \ckz Speedup vs official baseline, site-timed & official site (L1, L2, Quantization) & headline speedups (\S\S\ref{sec:boards}--\ref{sec:distribution}) \\
    Speedup vs official baseline, local & our B200s & search reward (\S\ref{sec:reward}), A/B tests (\S\ref{sec:design}) \\
    \ckz Speedup vs \texttt{torch.compile} & our B200s & local reference only (\S\ref{sec:distribution}) \\
    \bottomrule
  \end{tabularx}
\end{table}

\paragraph{Why Blackwell is hard.} Getting close to the bound often needs Blackwell-only features, which libraries do not yet cover for most fused or low-precision shapes. The site's own speed-of-light reference rows put a perfect kernel at about 10$\times$ the baseline on L1 and L2, 29$\times$ on FlashInfer-Bench and 38$\times$ on Quantization. That leaves a lot of room to gain, and a lot of room to go wrong.

\paragraph{Why correctness is hard.} A wrong kernel usually does not crash. It returns plausible numbers. Our agents' kernels have a median length of about 7,000 lines, more than a person can audit at the rate agents write them. A search strong enough to find fast kernels will also find gaps in any checker and learn reward hacks: ways to score without real work\footnote{The benchmark's authors report reward hacks in their own agent runs, in which 589 submissions (14.5\%) were flagged and rejected~\citep{lin2026solexecbench}.}.

\section{CompoundingKernel}
\label{sec:approach}

\subsection{The Child and Controller Loops}
\label{sec:archive}

In loop \circled{1} of Figure~\ref{fig:F2}, a child takes one parent (or none, for a new root) and one direction, and in the final run it stops at its first correct kernel. In loop \circled{2}, each problem's single controller has two tools: it queries the archive, and it calls \texttt{expand} with a parent and a list of distinct directions, one for each free child slot. One controller step is

\[
  (p, d) \sim \pi_{\mathrm{LLM}}(\cdot \mid A), \qquad k' = \mathrm{Child}(p, d), \qquad A \leftarrow A \cup \{k'\}\ \text{if } k' \text{ is correct}.
\]

Here $A$ is the problem's archive of correct kernels (\S\ref{sec:loops}). The controller LLM's choice given $A$, written $\pi_{\mathrm{LLM}}(\cdot \mid A)$, is a parent $p \in A \cup \{\bot\}$ ($\bot$ opens a new root) and a direction $d$. The child session $\mathrm{Child}(p, d)$ writes the kernel $k'$. The controller keeps no state of its own: it re-reads the archive each round, refills a slot as soon as a child ends, and runs until the remaining time that \texttt{expand} reports is used up. Loops \circled{1} and \circled{2} both run Claude Opus 5.5, which stays fixed throughout.

\subsection{Shared Memory Outside the Model}
\label{sec:loops}

\begin{figure}[!htbp]
  \centering
  \includegraphics[width=\linewidth,height=0.75\textheight,keepaspectratio]{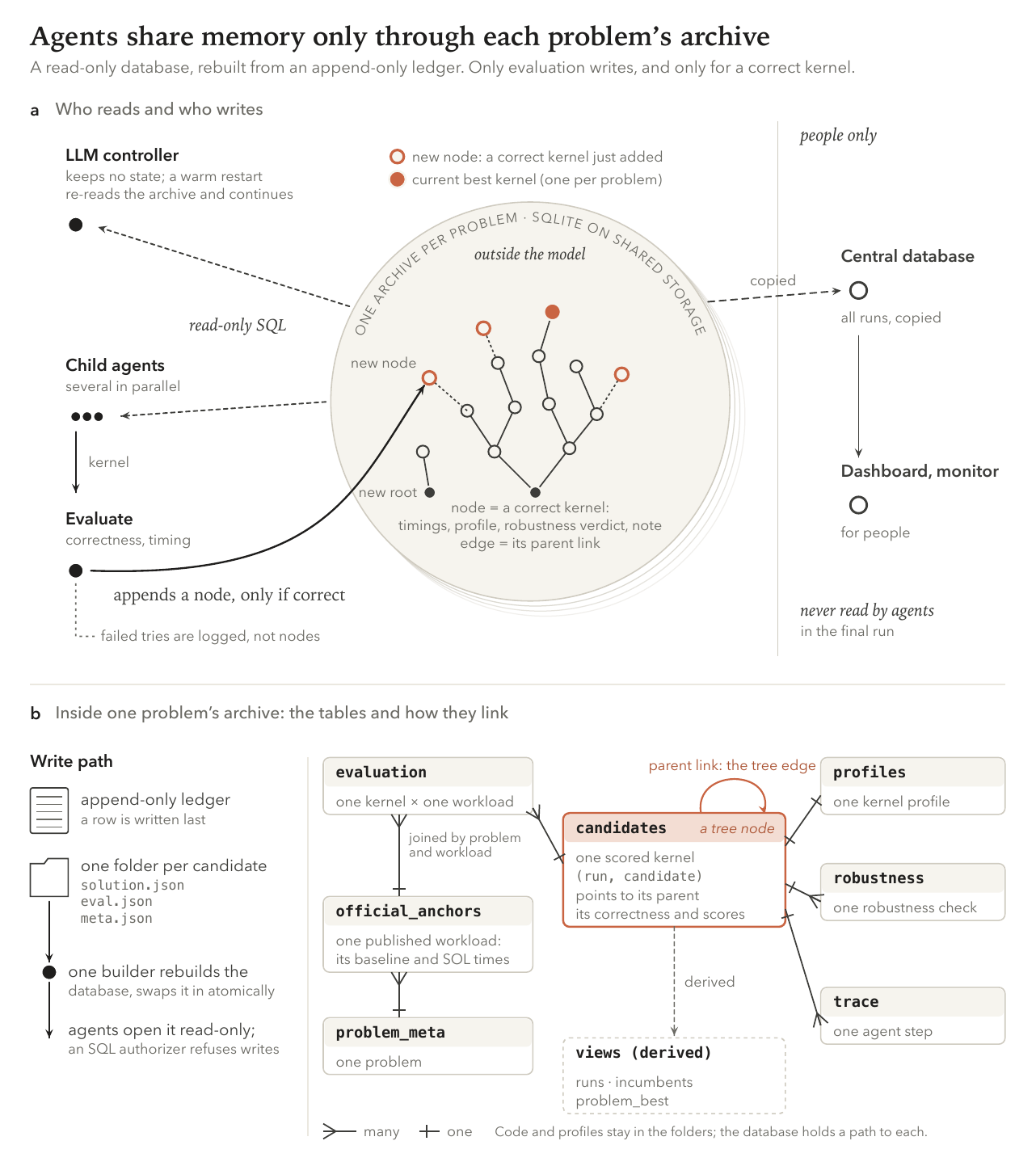}
  \caption{The archive is the agents' only shared memory: they read it with read-only SQL, and only evaluation appends a correct kernel. Most tables hang off one scored kernel, whose parent link is the tree edge.}
  \label{fig:F2b}
\end{figure}

A long search makes far more measured attempts than an agent can hold in context, so each run's memory lives outside the model, in an archive (Figure~\ref{fig:F2b}) and each run's archive is separate and unseen by other runs, so final runs will not be affected by our experiment runs. Figure~\ref{fig:F2b}a shows who reads and who writes. The controller and its children read the archive with read-only SQL. Only an evaluation appends a node, and only for a correct kernel; failed tries are logged but are not nodes. Agents keep no state of their own, so after a warm restart (within the same run) they simply re-read the archive. People's dashboard reads a separate copy which agents can not see. Figure~\ref{fig:F2b}b, left, shows the write path: each run appends to a ledger and writes one folder per candidate, a single builder rebuilds the SQLite database from the ledgers and swaps it in atomically, and agents open the result read-only, so they never see a half-written database. Figure~\ref{fig:F2b}b, right, shows the tables. Candidates sit at the center as tree nodes, and each one's parent link is a tree edge; a node without a parent is a root, meaning a fresh design. Per-workload evaluations join the published baseline and SOL times, and profiles, robustness checks and agent steps hang off each candidate. Appendix~\ref{sec:app-d} lists the tables.

\subsection{Slower Kernels as Stepping Stones}
\label{sec:stepping}

\circled{2} The controller may expand any node along the archive's parent links (Figure~\ref{fig:F2b}b), not only the current best. A correct but slower kernel can be a stepping stone: a new structure that is slower at first can lead to gains that small edits of the best kernel would not reach. The best path of Quantization/030 rose from 6.25$\times$ to 27.36$\times$ (local), yet 76 of its 270 steps were level or slower (\S\ref{sec:cases}).

\subsection{Reward from Local Timing}
\label{sec:reward}

For a correct kernel $k$, the reward is the geometric-mean speedup over the benchmark's published baseline, over the workloads $\mathcal{W}$ (\S\ref{sec:challenge}):

\[
  R(k) = \Bigl(\prod_{w \in \mathcal{W}} \frac{t_b(w)}{t_k(w)}\Bigr)^{1/|\mathcal{W}|}.
\]

\circled{1} Children measure $t_k(w)$ on our own B200s and store it in the per-workload evaluation table (Figure~\ref{fig:F2b}b), so no official score enters the reward. We did not use the SOL score as the reward because it flattens out at high speedups; agents see it only as context. Our GPUs run the SM clock unlocked, at up to about 1,965 MHz, so local times read fast. \circled{4} People correct for this with a calibration that agents never see (Appendix~\ref{sec:app-a}).

\subsection{Agents Never See Official Scores}
\label{sec:cleanroom}

The search loops \circled{1} and \circled{2} of the final run ran without people. Each final-run search started cold and could query only its own run's archive (Figure~\ref{fig:F2b}a). Official scores, submission results and our calibration never reached the agents. Agents could use the CUTLASS headers, the internet, and optional reference docs and skills, such as a public wiki.

People took part only in loops \circled{3} and \circled{4} (Table~\ref{tab:roles}). They built the early system, approved harness fixes at warm restarts, and decided what to send and when to publish, using the public standings.

Official scores never enter the search. No agent sees a score, a rank or a submission result, and the reward uses only our own timings. People used the public standings outside the search, in loop \circled{4}: to decide what to send and when, and to decide where compute goes. From 2026-10-25, humans set each problem's child slots from its gap to \#1: problems close behind \#1 got more, problems safely at \#1 gave some up, and the fleet total stayed the same. The agents received only a slot count per run. People also added a few approved notes to some runs' prompts (design directions, Triton guidance, a clock note); none of them contained a score or a rank.

\begin{table}[!htbp]
  \centering\cktablesetup\renewcommand{\arraystretch}{1}
  \caption{Who did what in the final run.}
  \label{tab:roles}
  \begin{tabularx}{\linewidth}{@{\hspace{4pt}}W{1}W{1}l@{\hspace{4pt}}}
    \toprule
    \thead{Activity} & \thead{Done by} & \thead{People} \\
    \midrule
    Kernel search; parents and directions & child and controller agents & none \\
    \ckz Harness fixes at warm restarts & lead and helper agents & approve \\
    Package and send & lead agent & approve each publish \\
    \ckz What to send, when; slot allocation & people & decide \\
    \bottomrule
  \end{tabularx}
\end{table}

\subsection{Checks Outside the Model Decide Correctness}
\label{sec:verify}

Every check that decides correctness runs outside the model, and its verdicts are stored with each candidate (Figure~\ref{fig:F2b}b). Table~\ref{tab:checks} groups the checks into five classes, and Appendix~\ref{sec:app-b} lists all twelve send checks.

\begin{table}[!htbp]
  \centering\cktablesetup\renewcommand{\arraystretch}{1}
  \caption{Classes of checks.}
  \label{tab:checks}
  \begin{tabularx}{\linewidth}{@{\hspace{4pt}}Q{3.1cm}W{1}W{1}@{\hspace{4pt}}}
    \toprule
    \thead{Class} & \thead{What it establishes} & \thead{When it runs} \\
    \midrule
    Official correctness & matches the reference on every workload & every evaluation \\
    \ckz Robustness & correct on unshipped shapes and stress values; memory-safe & each new best; multi-seed on each send \\
    Rules and reward hacks & speed and correctness come from real work & before a best counts; every send \\
    \ckz Site readiness & builds in time; passes a site rehearsal & every send \\
    Review and official score & checklist verdict, a person's go, the site's checks & every send \\
    \bottomrule
  \end{tabularx}
\end{table}

Three pieces of evidence show that the checks work. First, the search found reward hacks on its own. We caught them during the run and added new checks to the gate before anything was sent. Second, during the run, we took a robust kernel over a fast one: an L2/079 best passed every shape but overflowed FP16 on stress inputs, and we dropped it and accepted a fall from 1.62$\times$ to about 0.48$\times$ (local) at the time. Third, in \circled{4}, the site marks all 235 final kernels correct.

\section{Results}
\label{sec:results}

Board ranks, per-problem counts and speedups are as of \boarddate{}. Other kernel-derived numbers are as of \freezedate{}; line counts and source patterns as of 2026-10-31. The site scored and timed every L1, L2 and Quantization kernel on our public rows. FlashInfer-Bench (FI) kernels were verified correct on the site as private rows; we do not publish their results.

\subsection{\texorpdfstring{\#1 on L1, L2 and Quantization, All 235 Correct}{\#1 on L1, L2 and Quantization, All 235 Correct}}
\label{sec:boards}

\textbf{As of \boarddate{}, we rank \#1 on the public L1, L2 and Quantization boards. Every one of the 235 problems has a kernel that the site marks correct.} A collection board ranks teams by mean SOL score over the collection. A per-problem \#1 means our public row tops that problem's own board, which is a separate ranking (\S\ref{sec:challenge}). We first held \#1 on Quantization (2026-10-20), then on L1 (2026-10-27), then on L2 (2026-10-29), each time when we made that board's rows public. We have held \#1 on L1 since then, apart from a short loss on 2026-11-07. We held \#1 on both L1 and L2 from 2026-10-29 to 2026-11-07. Other teams moved ahead of us on L2 on 2026-11-07, and on Quantization on 2026-10-27 and 2026-11-08. Our public scores did not drop. We took back \#1 on L2 and Quantization when we made more of our rows public late on 2026-11-10 (UTC).

\begin{table}[!htbp]
  \cktablesetup\renewcommand{\arraystretch}{1.32}\setlength{\tabcolsep}{7.5pt}%
  \caption{\textbf{Official results by board.} Ranks and public SOL means come from the public boards at \boarddate{}, and per-problem counts from the per-problem boards at the same date. Board screenshots are in Appendix~\ref{sec:app-boards}.}
  \label{tab:boards}
  \begin{tabularx}{\linewidth}{@{\hspace{4pt}}L
      S[table-format=3.0]
      l
      S[table-format=1.6]
      S[table-format=3.0]
      S[table-format=3.0]
      S[table-format=1.2]@{\hspace{9pt}}}
    \toprule
      & & \tgroup{2}{Public board} & \tgroup{2}{Per-problem boards} & \\
    \cmidrule(lr){3-4}\cmidrule(lr){5-6}
    \multicolumn{1}{@{\hspace{4pt}}l}{\thead{Board}} & \theadr{Problems} & \thead{Our rank} & \theadr{SOL mean}
      & \theadr{\#1} & \theadr{Top 10} & \theadr{Speedup vs $t_b$ \tunit{(site, \texttimes)}} \\
    \midrule
         L1           & 94 & \#1          & 0.823339 & 35 & 84 & 4.04 \\
    \ckz L2           & 82 & \#1          & 0.819260 & 39 & 76 & 3.62 \\
         Quantization & 33 & \quantrank{} & 0.909670 & 15 & 33 & 9.19 \\
    \midrule
    \ckz \textbf{Public boards} & \bfseries 209 & \tna & {\tna} & \bfseries 89 & \bfseries 193 & \bfseries 4.40 \\
    \bottomrule
  \end{tabularx}
  \cknotes{Speedup is the geometric mean over problems of the site-timed speedup over the benchmark's baseline $t_b$, timed by the site on our public rows as of \boarddate{}. The table covers the three boards we submitted to.}
\end{table}

Our kernels are \textbf{4.40$\times$ faster than the benchmark's baseline over the 209 public-board problems} (geometric mean; median 3.78$\times$; site timing of our public rows). The board standing comes from breadth: we are \#1 on 89 of 209 problems and in the top 10 on 193 (Figure~\ref{fig:F1}).

\subsection{Speedups Are Broad, Not Driven by Outliers}
\label{sec:distribution}

\begin{figure}[!htbp]
  \centering
  \includegraphics[width=0.8\linewidth]{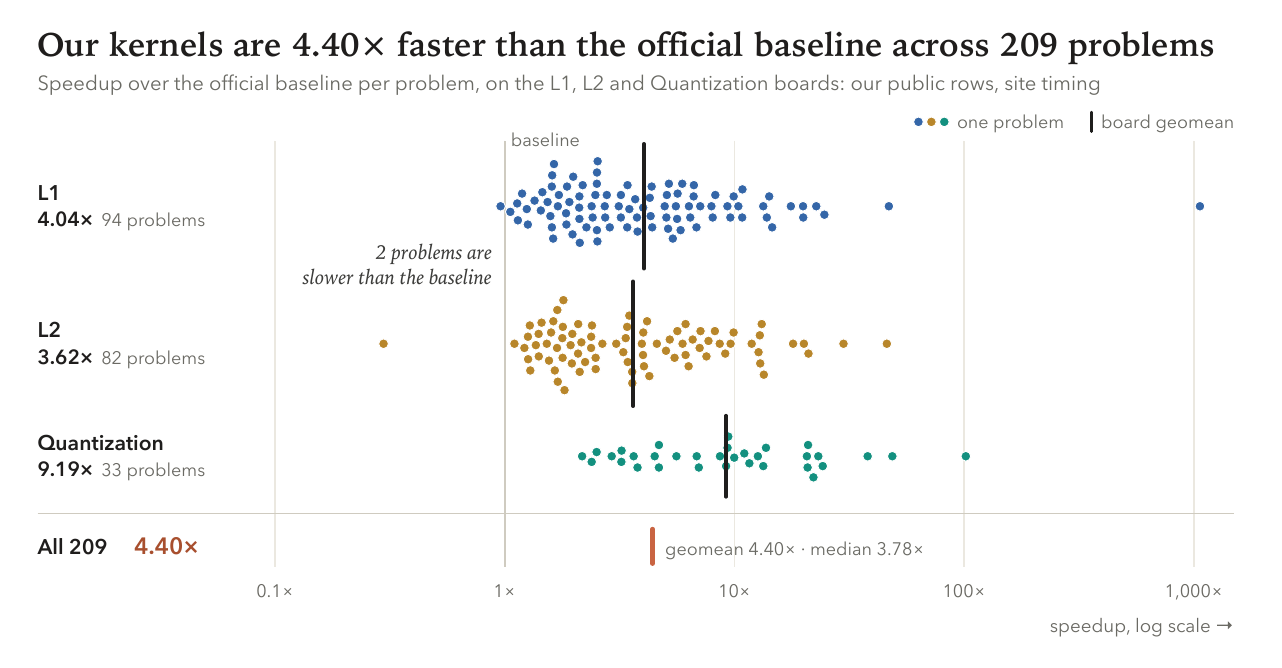}
  \caption{Our kernels are 4.40$\times$ faster than the official baseline over the 209 problems of the L1, L2 and Quantization boards (median 3.78$\times$). Each dot is one problem and each tick a board's geomean. Site timing of our public rows, as of \boarddate{}.}
  \label{fig:F3}
\end{figure}

\textbf{Most problems sit between 2$\times$ and 10$\times$. A handful of extreme speedups do not drive the mean.} Figure~\ref{fig:F3} plots one speedup per problem, on the same basis, as one beeswarm per board. The board geometric means are 4.04$\times$ (L1), 3.62$\times$ (L2) and 9.19$\times$ (Quantization). 39 problems run at least 10$\times$ faster than $t_b$, and 2 run at least 100$\times$ faster. The maximum is L1/094 at 1062$\times$. Two problems end slower than the baseline (\S\ref{sec:kernels}).

\emph{Local reference.} On our pods, the kernels as of 2026-11-10 are 11.2$\times$ faster than \texttt{torch.compile} (default mode). For the same kernels, the site-basis speedup was 4.47$\times$ on that date. The gap between the two has two causes. The pod clock reads fast (about 1.19$\times$), and \texttt{torch.compile} is a slower baseline than $t_b$ (about 2.11$\times$). The benchmark's baseline is the comparison that counts.

\subsection{Most Gains Came Early}
\label{sec:climb}

\begin{figure}[!htbp]
  \centering
  \includegraphics[width=0.8\linewidth]{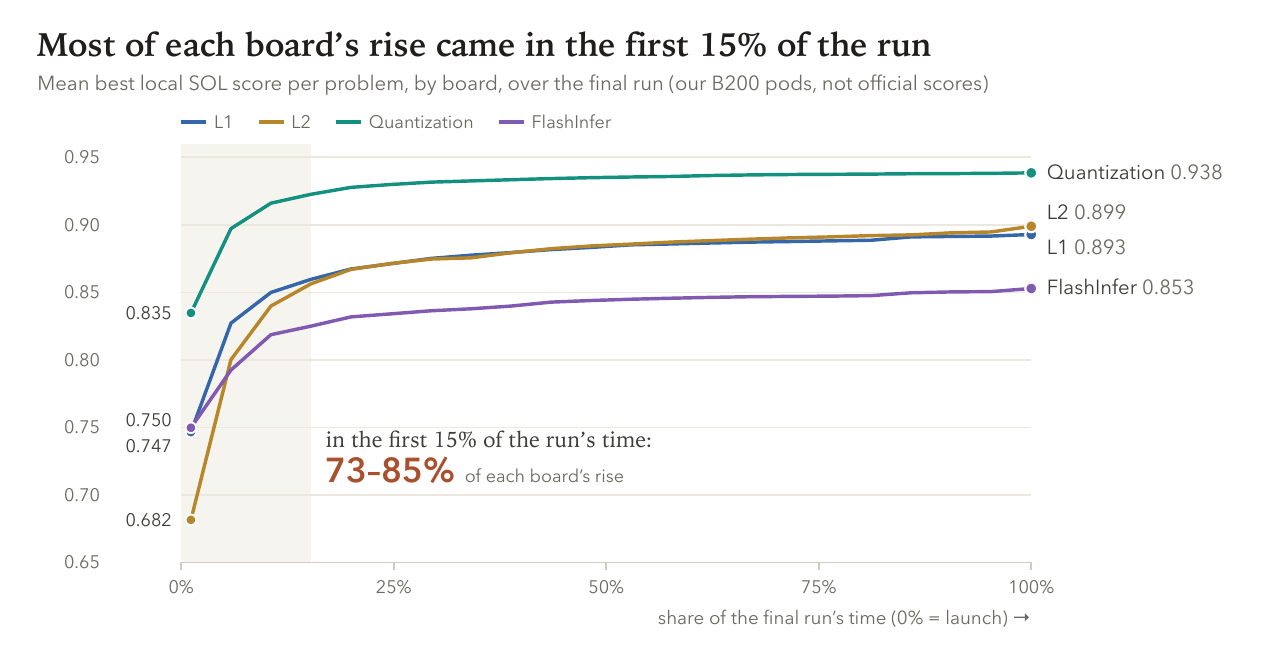}
  \caption{Most of each board's rise came in the first 15\% of the run: 73--85\% of the rise in mean best local SOL. Local scores on our B200 pods, not official scores.}
  \label{fig:F4}
\end{figure}

\textbf{Most of the gain came early in the run, and the search kept improving after that at a falling rate.} Figure~\ref{fig:F4} plots each board's climb as its mean best local SOL against the share of the run's time. From the first data point (about 1\% of the run's time) to the end of the data window, L1 rose from 0.747 to 0.893, L2 from 0.682 to 0.899, Quantization from 0.835 to 0.938 and FlashInfer from 0.750 to 0.853. In the first 15\% of the run's time, each board gained 73--85\% of its rise. Counted by evaluated candidates instead (Figure~\ref{fig:F4b} in Appendix~\ref{sec:app-f}; gain is a problem's best local SOL so far minus that of its first correct kernel, pooled over all 235 problems), the first 2\% of candidates delivered 81\% of the gain. The median problem gained 0.27 local SOL on L1, 0.39 on L2 and 0.25 on Quantization. The late tail is small in SOL, but it decided many per-problem ranks.

\textbf{Candidate counts follow difficulty, not payoff.}  Across problems, the rank correlation between a problem's correct candidates and its gain is $-$0.36 (Spearman, 235 problems). Hard problems with little room left absorbed many tries, and later in the run child slots moved toward problems close behind \#1 (Limitations, \S\ref{sec:conclusion}). The candidate axis therefore shows what the gain cost. It does not show that more tries on one problem buy more gain there.

\subsection{Blackwell-Native Kernels, and Where We Lag}
\label{sec:kernels}

\textbf{The final kernels are large, hand-level Blackwell code, not library wrappers.} By declared language, 214 of 235 are CUDA C++, 13 are CuTe DSL, 6 are Triton and 2 are mixed. All 78 per-problem \#1s of 2026-11-10 are CUDA C++ or CuTe DSL. 221 kernels exceed 1,000 lines, and the median is about 7,000. By a text match on the source, 77\% name \texttt{tcgen05} tensor-core instructions, 72\% issue TMA copies and 94\% contain inline PTX. None uses Hopper's \texttt{wgmma}. These counts come from searching the source text, so they show what the code contains, not what runs when the site times it.

\textbf{Where we fall short: plain GEMM/MLP work and full layers are furthest from SOL} (Table~\ref{tab:sol-by-type}). 16 of the 209 public problems sit outside the top 10 (10 on L1, 6 on L2). Two problems end slower than the baseline: L2/037 (0.30$\times$) and L1/030 (0.96$\times$). Of the 209 public problems, 21 are below 1.5$\times$. All 21 are on L1 or L2, with a mean site SOL of 0.61 against 0.86 for the rest. The worst case is partly self-imposed. On L2/037 a code review rejected a library fallback that had passed every automatic check (\S\ref{sec:verify}).

\section{Systems Lessons at Scale}
\label{sec:lessons}

These four lessons are about running over a thousand agent sessions at once. Each lesson follows challenge $\to$ intervention $\to$ measured impact $\to$ lesson. Most measurements are before/after readings inside the live run. Tests of the search itself follow in \S\ref{sec:design}.

\subsection{More Agents Does Not Mean More Experiments}
\label{sec:lesson-1}

\begin{figure}[!htbp]
  \centering
  \includegraphics[width=\linewidth]{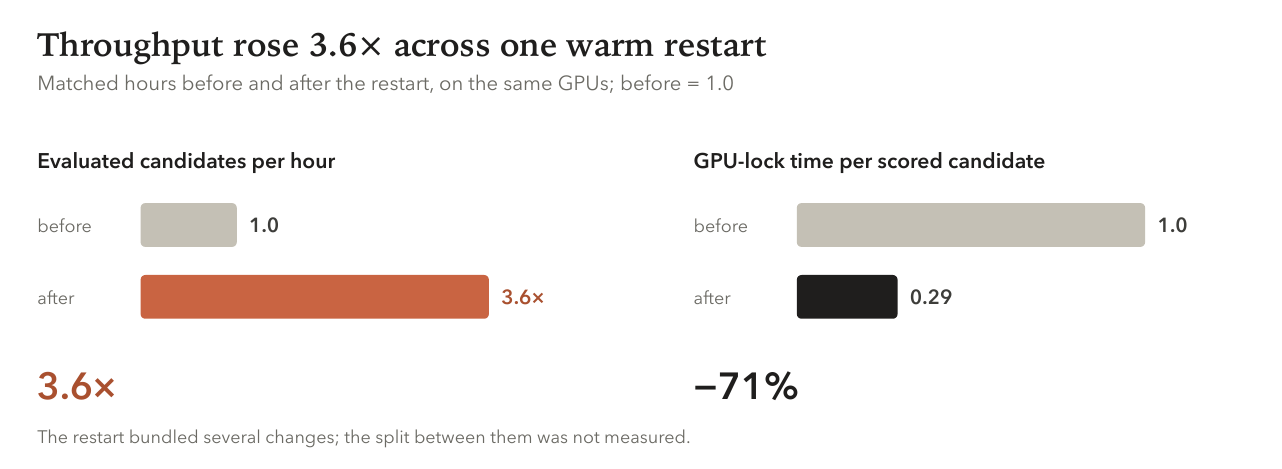}
  \caption{Throughput rose 3.6$\times$ across one warm restart, and GPU-lock time per scored candidate fell 71\%. Matched hours on the same GPUs, before = 1.0.}
  \label{fig:F5}
\end{figure}

\textbf{One warm restart raised evaluated candidates per hour 3.6$\times$. The lock waits point to the rate of reliable measurements, not the number of agents, as the limit.}
\ckri{challenge}{Challenge}Early on, all GPU locks were held and about five calls per GPU were waiting, yet the GPUs were only about 13\% busy. Each lock covered the whole remote call, including the CPU build and a 12--15 s shell start. More children per problem would likely have added waiters. \ckri{intervention}{Intervention}One early warm restart bundled several changes: builds moved out of the lock, shell start fell to milliseconds, profiling became optional, and the infrastructure took over validating new bests. \ckri{measured}{Measured}In matched hours before and after, evaluated candidates per hour rose 3.6$\times$ and lock time per landed candidate fell 71\% (same GPUs and slots; Figure~\ref{fig:F5}). The correct share stayed at about 98\%. The 3.6$\times$ belongs to the whole restart plus other progress between the two windows. A second restart moved the lock to the host and doubled the hosts. That more than doubled the hourly median again, but it mixes two changes. We have not yet tested the limit directly by varying the number of agents.

\begin{cklesson}\ckri{lesson}{Lesson}Agent count measures demand. What made tries useful was reliable, fast feedback on each one: a correct measurement that the next attempt could build on. The same rate explains the controller's win in \S\ref{sec:lesson-5}.\end{cklesson}

\subsection{Warm Restarts Keep Progress}
\label{sec:lesson-2}

\textbf{The final run took its fixes through more than a dozen warm restarts and never started over.}
\ckri{challenge}{Challenge}A long run must absorb fixes, checks and hardware without losing progress. \ckri{intervention}{Intervention}Search memory lives outside the agents, in each run's archive and ledger. A warm restart applies one change and relaunches the agents on the same run. Each controller then re-reads its archive and continues from its best kernels. After one restart kept the agents down for hours, we adopted a relaunch kit and a rule: one gate-able change per launch. \ckri{measured}{Measured}Agents-only relaunches later cost a few minutes of agent downtime. 

\begin{cklesson}\ckri{lesson}{Lesson}Memory outside the model makes the harness something you can iterate on. ``The method'' then becomes an ordered sequence of versions, so we publish the sequence.\end{cklesson}

\subsection{Verify What You Might Ship}
\label{sec:lesson-3}

\textbf{Checking new bests and sent rows, instead of every candidate, cut robustness checks per landed candidate from 0.66 to 0.27.}
\ckri{challenge}{Challenge}When children ran the robustness check themselves, at about 0.77 checks per candidate, it took an estimated quarter of lock time while the lock was the bottleneck. \ckri{intervention}{Intervention}Plain code checks each run's new best automatically, and every sent row gets the full check. Other candidates still pass the local correctness test. \ckri{measured}{Measured}We compared before and after, inside the first warm restart. About 95\% of new bests were still checked, within minutes of landing, and 95.9\% of those were robust.

\begin{cklesson}\ckri{lesson}{Lesson}Put verification where the decision is, and measure the coverage you give up.\end{cklesson}

\subsection{Pass Parents as Files, Not Text}
\label{sec:lesson-4}

\textbf{Passing large parents as files cut the context written by first prompts 88--96\%.}
\ckri{challenge}{Challenge}Each child received its parent's full solution in its first prompt. Pasted parents made up most of the first prompt's context. \ckri{intervention}{Intervention}In a late restart, large parents arrive as files with only their paths in the prompt. Three smaller changes shipped in the same restart. \ckri{measured}{Measured}In matched windows before and after, candidates per hour stayed about the same.

\begin{cklesson}\ckri{lesson}{Lesson}With over a thousand agents running, context size matters across the whole fleet. Measure what fills the context. Here the change kept the same throughput with far less context written by first prompts.\end{cklesson}

\section{Search-Design Findings}
\label{sec:design}

Three controlled comparisons test the search itself (Table~\ref{tab:design}): an LLM controller against a coded UCB style tree, mandatory against optional reference docs and skills, and structural steering of stuck runs. All three ran on pre-final harness versions or as short screens, with one run per arm; we did not repeat them on the final harness. Appendix~\ref{sec:app-e} gives each test in full.

\begin{table}[!htbp]
  \cktabledense\renewcommand{\arraystretch}{1}
  \caption{\textbf{Design comparisons.} All rows compare the official $t_b$ against local kernel timings. Phase 1 and Phase 2 are pre-final harness versions (Appendix~\ref{sec:app-e}).}
  \label{tab:design}
  \begin{tabularx}{\linewidth}{@{\hspace{4pt}}W{1.15}>{\raggedleft\arraybackslash\rule{0pt}{10.5pt}}p{1.2cm}<{\rule[-4.5pt]{0pt}{0pt}}W{0.74}W{1.13}W{0.98}@{\hspace{4pt}}}
    \toprule
    \thead{Comparison} & \thead{n} & \thead{Horizon} & \thead{Effect} & \thead{Verdict} \\
    \midrule
         LLM controller vs coded UCB style tree & 38 pairs & 24 h per arm\tsub{Phase 1}
           & better on 35 of 38 at equal time;\newline better on 15 of 38 at equal count & wins by 2.4$\times$ correct attempts \\
    \ckz Mandatory vs optional reference docs and skills & 233 pairs & $\sim$23 h\tsub{Phase 2}
           & ratio 0.981;\newline control ahead 126 vs 107 & no measured lift \\
         Structural steering of stuck runs & 4 runs & 210 min\tsub{final run}
           & $-$0.30\% to +0.05\% & screening null; not shipped \\
    \bottomrule
  \end{tabularx}
  \cknotes{Appendix~\ref{sec:app-e} gives every test in full, with its p-values and confounds.}
\end{table}

\subsection{The Controller Wins by Making More Attempts}
\label{sec:lesson-5}

\begin{figure}[!htbp]
  \centering
  \includegraphics[width=0.85\linewidth]{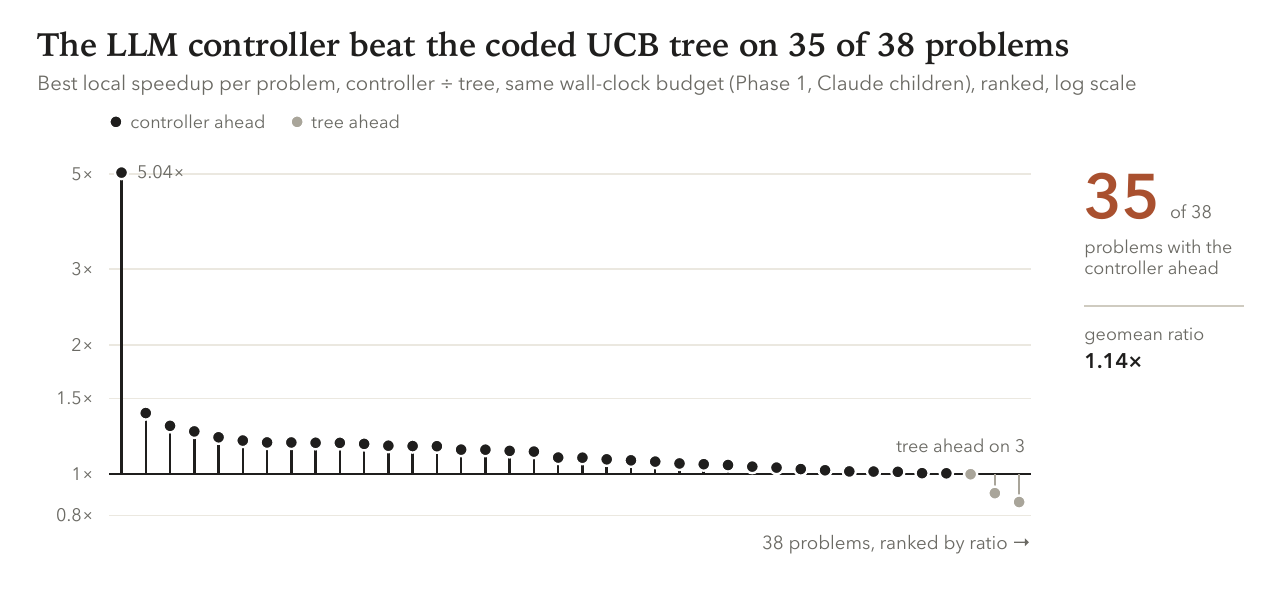}
  \caption{At an equal wall-clock budget, the LLM controller beat the coded UCB style tree on 35 of 38 problems (34 at a strict 24 h cut). Each stem is one problem's best local speedup ratio, log scale.}
  \label{fig:F6}
\end{figure}

\textbf{At equal wall time, the LLM controller beat a coded UCB style tree on 35 of 38 paired problems. It won by making 2.4$\times$ as many correct attempts, not by better picks.} Both arms used the same Claude child agents (the tree's children got no direction) and the same archive design, with one archive per arm. Each had a 24 h budget on the Phase 1 harness, with 40 problems across the four boards (Figure~\ref{fig:F6}). The controller was better on 34 of 38 if both arms are cut exactly 24 h after launch. At an equal candidate count, the arms are about even. Over its first candidates, as many as the tree made, the controller did better on 15 problems and worse on 23 (16 and 22 against a locally timed baseline). Per attempt, the controller picks about as well as the tree. The lead comes from making more attempts in the same time: full batches every round with LLM-chosen directions, and no width narrowing (Appendix~\ref{sec:app-e1}). This is the limit of \S\ref{sec:lesson-1} again.

\subsection{Added Guidance Gave No Measured Lift}
\label{sec:design-guidance}

\begin{figure}[!htbp]
  \centering
  \includegraphics[width=0.85\linewidth]{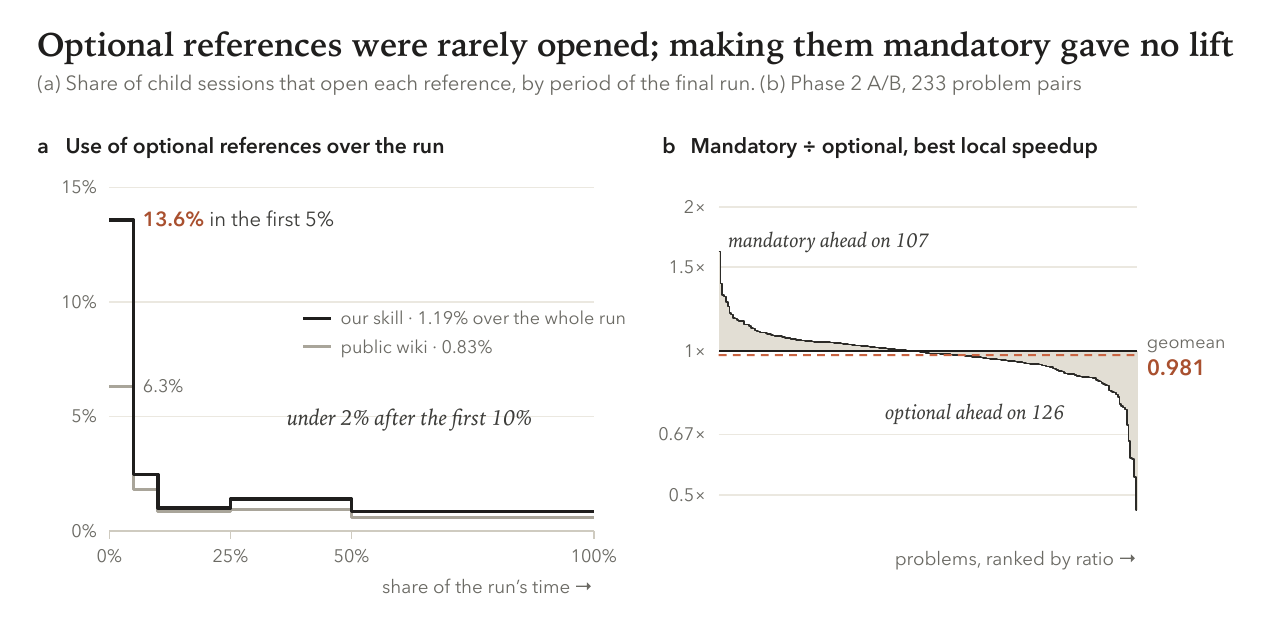}
  \caption{Optional references were rarely opened, and making them mandatory gave no lift. (a) Share of child sessions opening each reference, 13.6\% at first and under 2\% after the first 10\%; (b) 233 A/B pairs, geomean ratio 0.981.}
  \label{fig:A2}
\end{figure}

\textbf{Making the optional reference docs and skills mandatory raised their use 4.3$\times$ but not the final best speedups.} On the Phase 2 harness experiments, both arms ran all 235 problems at the same time, each on half of the fleet, for about 23 h. Over 233 pairs, the geomean ratio of gated to control was 0.981 (Figure~\ref{fig:A2}). When the references were optional, their use in the final run fell from 13.6\% of child sessions (our skill) in the first 5\% of the run's time to under 2\% after the first 10\%. Over the whole final run, 1.19\% of child sessions opened our skill and 0.83\% the public wiki. A structural-change instruction for stuck runs also showed no effect in a screening test, and we did not ship it. The result agrees with goal-only prompting~\citep{chen2026justgoal} and contrasts with reports that harness-supplied knowledge helps~\citep{shui2026harness}. Two confounds worked against the gated arm (Appendix~\ref{sec:app-e3}).

\begin{cklesson}\ckri{lesson}{Lesson}Under these conditions, more reliable attempts per hour paid off measurably, and added guidance did not.\end{cklesson}

\nobarriersection
\section{Case Studies: Slower Kernels Led to the Best}
\label{sec:cases}
\restorebarrier

\begin{figure}[!htbp]
  \centering
  \includegraphics[width=\linewidth]{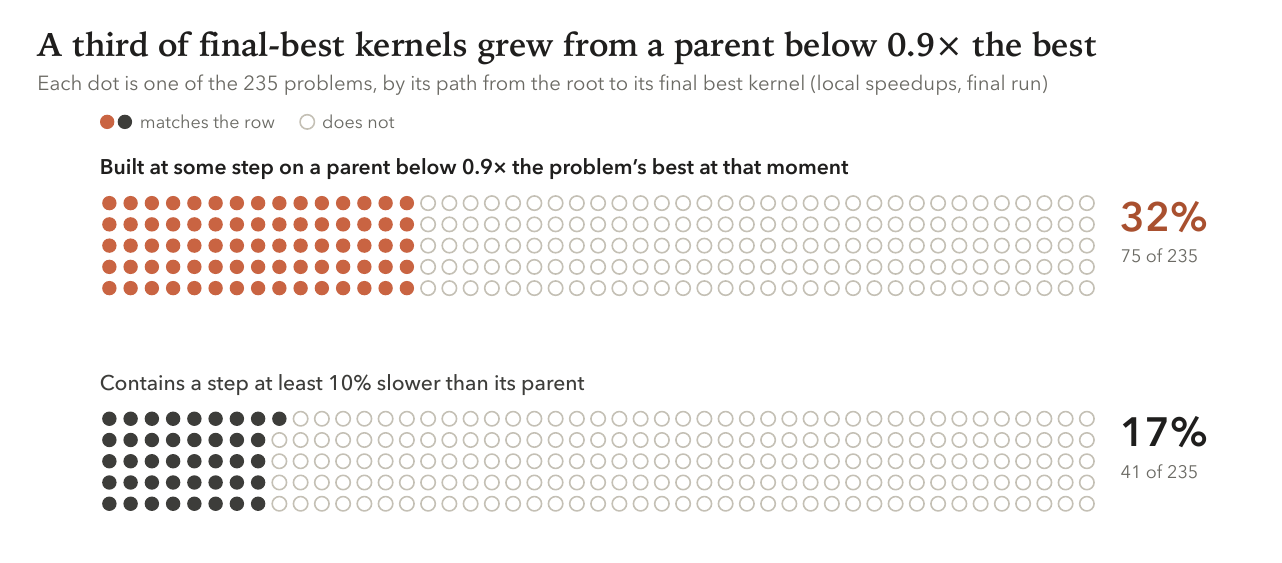}
  \caption{A third of final-best kernels (75 of 235) were built at some step on a parent below 0.9$\times$ the problem's best at that moment. Each dot is one problem; local speedups, final run to \freezedate{}.}
  \label{fig:F8}
\end{figure}

Many final kernels descend from tries that were slower than predecessors. Across the 235 final-run problems, 41 best paths (17\%) contain a step that lost at least 10\% of local speed against its parent, and on 75 (32\%) paths, the search at least once extended a node that was below 0.9$\times$ the run's best at that moment (local timing, as of \freezedate{}; Figure~\ref{fig:F8}). Four problems below show why keeping such nodes paid off; a fifth shows a fast local kernel that failed the site's build. We favoured cases whose sent kernel is robust, public and free of runtime timing. An archive edge records only the seed: a child may also copy code from other nodes, and three cases turn on such a graft.

\begin{figure}[!htbp]
  \centering
  \includegraphics[width=\linewidth]{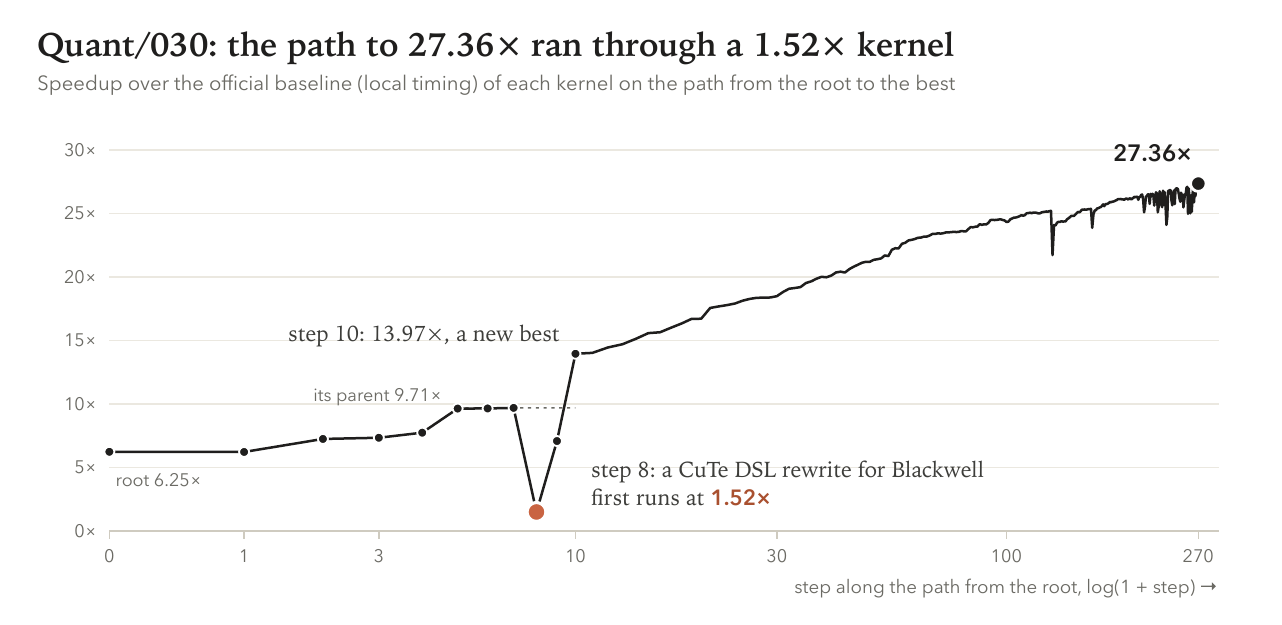}
  \caption{Quant/030's path to 27.36$\times$ ran through a 1.52$\times$ kernel: a CuTe DSL rewrite first ran slowly, then reached 13.97$\times$ two steps later. Local timing against the official baseline; the x-axis is log(1 + step).}
  \label{fig:F7}
\end{figure}

\paragraph{Quant/030: a slower attention kernel became the base of the best path.} The problem is an NVFP4 decoder layer (pre-norm, QKV projection, attention, output projection). Its best path rose from 6.25$\times$ to 27.36$\times$ over the official baseline (local, at the case-study cut; Figure~\ref{fig:F7}). Its deepest fall came just before one of its largest gains. At 9.71$\times$, the controller directed a child to replace the Triton attention with a CuTe DSL attention on Blackwell's tensor-core pipeline (tcgen05 MMA into tensor memory). The first try ran at 1.52$\times$. The next child traced the slowdown to register spills and reached 7.10$\times$, still below the parent. A third child moved the new attention onto the run's best node at the time (12.93$\times$, library NVFP4 GEMMs) and reached 13.97$\times$, a new run best about 1\% of the run's time after the dip. 93\% of the run's nodes descend from the 1.52$\times$ node; the best node not descended from it reached 24.87$\times$ (local, at the case-study cut).

A later episode needs stating plainly. From path step 3 to step 155 (roughly the first third of the run), kernels on this path chose between variants per shape with a short timing run on first use (first an autotuner, later a timing run when the launch plan was built). This reward hack was caught and our rules banned it later. When the ban came, a child removed every such run and fixed the choices in a per-shape table. This cost 6\% locally (25.36$\times$ $\to$ 23.88$\times$), and the path was back to 25.42$\times$ about 1\% of the run's time later. The sent kernel (site SOL 0.971073, 20.67$\times$ site-timed, faster than the baseline on 16 of 16 workloads) and the later 27.36$\times$ best both descend from that timer-free node.

\paragraph{L2/001: a step down for correctness.} This is the backward pass of a vision attention block with layer norms, in fp32 with a tight tolerance. The fastest kernel early in the run (5.73$\times$, local) used single-rounding fp16 tensor-core math in its projection GEMMs. It passed the official workloads, but our robustness check with more shape tests marked it NOT\_ROBUST (failed some extra shapes). On those shapes, the rounding error grows with the reduction length. Its child switched those GEMMs to split-precision operands and fell to 4.26$\times$. A second fix gave 4.17$\times$, and that node passed all six shapes. The search built on it, passed the old 5.73$\times$ about 3\% of the run's time after the fix, and reached 15.20$\times$, 2.65$\times$ the fragile kernel. 86\% of its nodes descend from the fix. The sent kernel scores SOL 0.936556 on the site (11.88$\times$ site-timed).

\paragraph{L2/043 and L1/065: lines far behind the leader.} L2/043 is a Mamba-2 chunked state-space scan. At about 11\% of the run's time, while the best was a multi-kernel chain at 32.08$\times$ (local), a new root appeared: a single fused, persistent tensor-core kernel at 15.10$\times$, under half the leader's speed. Soon after, a child kept the leader's chain for most shapes and used the fused kernel for the one shape class where it was faster. The result, 32.15$\times$, was a new run best; the line went on to hold 97\% of later nodes and reach 42.86$\times$ (1.34$\times$ the earlier leader). Its sent kernel scores SOL 0.971494 on the site (29.83$\times$ site-timed). L1/065, a fused 1$\times$1 projection with a split output, shows the same pattern on the L1 board. A tf32 tensor-core line sat at 1.40$\times$ while a small-shape kernel on a sibling branch led at 1.93$\times$. Soon after, a child sent small shapes to the leader's kernel and large shapes to the tensor-core kernel. That hybrid reached 2.10$\times$ and climbed to 5.63$\times$, and the line holds 96\% of later nodes (site SOL 0.831881). In both, the lagging node supplied a part that won on some shapes, and grafting it onto the incumbent made the new base.

\paragraph{Quant/029: fast locally, rejected by the site's build.} Quant/029 is an NVFP4 fused MLP (gate-and-up GEMM, SiLU gating, down GEMM). Its local best reached 14.55$\times$. However, it compiled every variant it might pick (41 to 71 source files) and chose among them by timing at run time. Three sends in a row failed the site's 300-second build limit; none failed on correctness. Our pods had built two of them in under three minutes. Built serially from a cold cache, as the site appears to build, the accepted kernels finished in at most 308 s and the refused ones took at least 1,965 s. A ``unity'' build of another candidate, which folds its many files into a few, was accepted (SOL 0.92697) and raised our private Quant score from 0.873267 to 0.901419. However, that row still selected variants by runtime timing, which was caught by our agent, so a timer-free kernel later replaced it, which deleted the configurations the plan can never reach (12 of 68 kept, four files). It was accepted at SOL 0.931935 (11.03$\times$ site-timed). Every send now passes a build-time check, which has stopped a newer 106-file candidate; the replacement row remains our Quant/029 entry.

In these cases, a node's local speed was a poor guide to its value as a parent: the dips in Quant/030 and L2/001 and the lagging lines in L2/043 and L1/065 each became the base of 86--97\% of later nodes. And a local score is a claim about our pods, not the site: on Quant/029, the score came from making the kernel smaller, not faster.

\section{Related Work}
\label{sec:related}

\begin{table}[!t]
  \centering\cktablesetup\renewcommand{\arraystretch}{1}
  \newcommand{\hw}[1]{\newline{\color{ckmuted}\fontsize{7.5}{9}\selectfont #1}}
  \caption{\textbf{Recent kernel systems with B200 or SOL-ExecBench results nearly all keep the model fixed and differ mainly in search structure and validation.} Entries are as stated in each work's own report; the grey line gives hardware and benchmark. Appendix~\ref{sec:app-i} lists 25 systems.}
  \label{tab:systems}
  \begin{tabularx}{\linewidth}{@{\hspace{4pt}}Q{3.9cm}W{1.1}W{0.9}@{\hspace{4pt}}}
    \toprule
    \thead{System} & \thead{Search} & \thead{Verification} \\
    \midrule
    \textbf{ForgeKernel}~\citep{chen2026forgekernel}\hw{H100, B200, Ascend; three suites} & architect plans milestones, coder implements; accepted kernels seed the next run & tolerance tests; no delegation to the reference \\
    \ckz \textbf{Cursor $\times$ NVIDIA}~\citep{cursor2026multiagentkernels}\hw{27 B200s; all 235 SOL-ExecBench} & one planner, many workers, shared markdown file & benchmark's SOL-violation check \\
    \textbf{Recursive}~\citep{recursive2026firststeps}\hw{B200; all 235 SOL-ExecBench} & long-horizon research threads, cross-task transfer & iterated hack detector; manual and AI review \\
    \ckz \textbf{KernelArc}~\citep{kundu2026kernelarc}\hw{H100, B200; SOL subset} & strategy-specialized parallel agents, shared memory & deterministic benchmark guard \\
    \textbf{DuoDiscover}~\citep{agarwal2026duodiscover}\hw{GPU and TPU suites; all 235} & agent fleet hill-climbs proxy evaluators & proxy fidelity tests \\
    \ckz \textbf{JustGoal}~\citep{chen2026justgoal}\hw{B200; 10 operators vs.\ libraries} & one goal-only agent & no looking at or copying the reference \\
    \textbf{Kevin-32B}~\citep{baronio2025kevin}\hw{RL-trained; H100/H200} & parallel trajectories $\times$ serial refinement & zero reward for PyTorch use \\
    \bottomrule
  \end{tabularx}
  \cknotes{\emph{Three suites:} expert libraries, SOL-ExecBench FlashInfer-Bench and CANN-Bench. Kevin-32B uses KernelBench L1--2.}
\end{table}

\paragraph{Kernel benchmarks and their integrity.} KernelBench~\citep{ouyang2025kernelbench} and TritonBench~\citep{li2025tritonbench} made LLM kernel generation measurable as speedup over a reference implementation, and such speedups proved easy to inflate. An early agentic result was traced to reused reference memory~\citep{miru2025sakanahack}, METR filtered cheatable tasks~\citep{metr2025kernel}, hardened re-evaluation moves the best KernelBench model from $1.43\times$ to $0.88\times$~\citep{zhang2026kernelbenchverified}, and standard checkers miss real faults~\citep{du2026measuringchecker,lange2025robustkbench}, and kernels often hardcode the tested shapes~\citep{younesian2026agentkernelarena}. SOL-ExecBench~\citep{lin2026solexecbench} scores each kernel against a speed-of-light bound from SOLAR~\citep{nvlabs2026solar} and rejected 14.5\% of its own agent submissions as reward hacks. In our run, agents never see an official score, and the public board runs and scores every submission itself~\citep{solexecbench2026leaderboard}.

\paragraph{Changing the model versus searching with a fixed one.} One line trains kernel models with RL or fine-tuning~\citep{baronio2025kevin,li2025cudal1,li2025autotriton,woo2025tritonrl,dai2026cudaagent,tehrani2026gpt5kernel,du2026kernelsmith,yuksekgonul2026tttdiscover}, mostly evaluated on KernelBench. The other keeps the model fixed and builds the loop around it~\citep{wang2025geak,wei2025astra,dong2025stark,zhang2025cudaforge,chen2025nvidiar1kernels,wang2025kernelfalcon}, increasingly with explicit search state~\citep{cao2026ksearch,ye2026cake,chen2026avo}. On Blackwell, the results that reach expert libraries or the top of public boards come from this second line~\citep{chen2026avo,nvlabs2026kda,chen2026forgekernel,kundu2026kernelarc,agarwal2026duodiscover,cursor2026multiagentkernels,recursive2026firststeps}. We belong to it (Table~\ref{tab:systems}).

\paragraph{How much the harness should know.} JustGoal finds that a goal-only agent beats open-source libraries on most of its operators, with no consistent benefit from domain knowledge or fixed workflows~\citep{chen2026justgoal}. A FlashInfer contest report finds expert directions essential~\citep{shui2026harness}, and AKO keeps a person to redirect stalled searches~\citep{luo2026ako}. Our mandatory-reference A/B (\S\ref{sec:design-guidance}) sides with JustGoal, on an early harness.

\paragraph{Archives, controllers and test-time compute.} Program-evolution systems keep an archive and sample parents from it~\citep{romeraparedes2024funsearch,novikov2025alphaevolve,openevolve,lange2025shinkaevolve,zhang2025dgm}, and strong results often descend from weaker intermediates~\citep{lehman2011abandoning,wang2019poet,hughes2024openendedness,cao2026ksearch}, and diverse tree search helps escape local optima~\citep{du2026adaexplore}. We adopted three ideas. The Darwin G\"{o}del Machine grows an archive by sampling a parent~\citep{zhang2025dgm}; we keep one such archive per problem, so many lineages grow in parallel. From Hyperagents we took an LLM meta agent above the task agents~\citep{zhang2026hyperagents}; in our system it also runs the search. From ForgeKernel we adapted the idea that a new root opened when a search stalls should be a new design~\citep{chen2026forgekernel}. Other work also moves search control into a model~\citep{liu2026evox,zheng2026dreamrsi,cao2026eat} or add progress-aware rules for backtracking and crossover~\citep{yan2026pacevolve}. Most scaling work adds samples or agents~\citep{brown2024monkeys,snell2024scaling,wang2026parallelteams}; a few treat evaluation as the bottleneck~\citep{agarwal2026duodiscover} or cut the cost of each trial~\citep{guo2026specgen,hari2026ucutlass}, as we found: adding agents did not add experiments until build and timing contention was removed.

\nobarriersection
\section{Limitations, Artifacts, and Conclusion}
\label{sec:conclusion}
\restorebarrier

\paragraph{What the work shows, and what it does not.} It shows that a fixed model, with official scores kept out of the search, can go from PyTorch definitions to a site-correct kernel for all 235 problems and to the board ranks of \S\ref{sec:results}. In our system, throughput rose 3.6$\times$ when one warm restart took builds out of the GPU lock, which points to the rate of reliable measurements, not agent count, as the limit on scale; we have not yet tested this by varying the agent count (\S\ref{sec:lesson-1}). The controller test points the same way: the LLM controller beat a coded tree on 35 of 38 paired problems at equal wall time by making 2.4$\times$ as many correct attempts (\S\ref{sec:lesson-5}). It does not show that our kernels are correct on all inputs, or that the method transfers to other hardware or benchmarks.

\paragraph{Limitations.} The evidence comes from one benchmark (SOL-ExecBench on B200), one model (Claude Opus 5.5 as every agent) and one cold-started final run, plus separate A/B runs. Correctness is tested, not proved: it rests on testing outside the model (\S\ref{sec:verify}, Appendix~\ref{sec:app-b}), because the formal and SMT-based tools we examined do not model Blackwell's asynchronous tensor-core operations (\texttt{tcgen05}, TMA, \texttt{mbarrier}). During the run the robustness check covered each run's new bests (94.9\% of them were checked), not every candidate (\S\ref{sec:lesson-3}). Local timing reads high against the site's (\S\ref{sec:reward}); people correct for this when choosing sends, and the estimate can miss when a design changes a lot (Appendix~\ref{sec:app-a}).

The cold start has the qualifiers stated in \S\ref{sec:cleanroom}. People chose which kernels to send and when to publish, using the public standings, and child slots followed each problem's gap to \#1. The design tests ran on pre-final harness versions or as a short screen, with one run per arm (\S\ref{sec:design}), so their verdicts may not hold for the final harness or the full run's horizon. Finally, the final run kept going after the data were read, so each headline number carries its as-of time in Appendix~\ref{sec:app-j}.

\paragraph{Artifacts.} We plan to release our code (the search loop, the SOL-ExecBench interface, the checks and an example configuration; not the agent prompts) and our final kernels for all 235 problems at \href{\demourl}{URL}, which already lists the 235 problems by board and operation type.

\paragraph{Conclusion.} Starting from PyTorch definitions, with a fixed model and official scores kept out of the search, CompoundingKernel wrote a site-correct kernel for every one of the 235 SOL-ExecBench problems. As of \boarddate{} it ranks \#1 on L1, L2 and Quantization. The gains came from making many measured tries build on each other, and the rate of reliable measurements, not agent count, appears to be the limit on scale. Every candidate still needs a full build and a timed run on a B200.

\paragraph{Next steps.}\phantomsection\label{sec:e2e} We will work on cheaper verification, so that more tries can be measured reliably, and test the loop on other GPUs and benchmarks. We will also swap our kernels into a serving framework for models with matching shapes, and measure time to first token and throughput. That gain will be smaller than the per-kernel speedups, because serving stacks already use tuned library kernels.

\FloatBarrier
\label{end-of-main-text}\makeatletter\pdfsavepos\write\@auxout{\gdef\string\endmainpos{page \thepage, y=\the\pdflastypos sp}}\makeatother

\bibliographystyle{unsrtnat}
\bibliography{references,references-v6}

\FloatBarrier
\appendix
\counterwithin*{table}{section}
\renewcommand{\thetable}{\thesection\arabic{table}}
\renewcommand{\thealgorithm}{C\arabic{algorithm}}

\section*{Notation}
\phantomsection\addcontentsline{toc}{section}{Notation}\label{sec:notation}

{\centering\cktablesetup\renewcommand{\arraystretch}{1}
\begin{tabularx}{\linewidth}{@{\hspace{4pt}}Q{3.3cm}W{1}Q{3.1cm}@{\hspace{4pt}}}
  \toprule
  \thead{Symbol} & \thead{Meaning} & \thead{Defined in} \\
  \midrule
  $w$, $\mathcal{W}$ & a workload (one input shape); the problem's set of workloads & \S\ref{sec:challenge} \\
  \ckz $k$, $t_k(w)$ & a kernel; its measured time on workload $w$ & \S\ref{sec:challenge} \\
  $t_b(w)$, $t_{\mathrm{sol}}(w)$ & published time of the benchmark's baseline on $w$; speed-of-light bound on $w$ & \S\ref{sec:challenge} \\
  \ckz $S_w(k)$ & SOL score of kernel $k$ on workload $w$ & \S\ref{sec:challenge}, Appendix~\ref{sec:app-a} \\
  $R(k)$ & reward: geometric-mean speedup of $k$ over $t_b$ on $\mathcal{W}$ & \S\ref{sec:reward}, Appendix~\ref{sec:app-a} \\
  \ckz $A$ & the problem's archive of correct kernels (one per run) & \S\ref{sec:archive}, \S\ref{sec:loops} \\
  $p$, $\bot$ & a parent node; ``none'' (as a parent, it opens a new root) & \S\ref{sec:archive} \\
  \ckz $d$, $D$ & one direction for a child; a list of directions & \S\ref{sec:archive}, Algorithm~\ref{alg:c2} \\
  $\pi_{\mathrm{LLM}}$ (written $\pi$ in Algorithm~\ref{alg:c2}) & the controller LLM's choice given the archive & \S\ref{sec:archive} \\
  \ckz $\mathrm{Child}(p, d)$, $k'$ & a child session started from $p$ with direction $d$; the kernel it writes & \S\ref{sec:archive} \\
  $P$, $r$ & a problem; a run & Appendix~\ref{sec:app-c} \\
  \ckz $n$, $\mathrm{spd}(n)$ & an archive node; its measured speedup over $t_b$ & Appendix~\ref{sec:app-c} \\
  $N$ & a child's cap on correct candidates (1 in the final run) & Appendix~\ref{sec:app-c} \\
  \ckz $B$, $T$, $\mathit{slots}$ & backstop budget; deadline; child slots & Algorithm~\ref{alg:c2} \\
  $\mathrm{UCB}(n)$, $c$ & the tree arm's node score; its constant (0.7) & Appendix~\ref{sec:app-c} \\
  \ckz $\mathit{best}$, $\mathrm{visits}(n)$ & top speedup among live nodes; number of children of $n$ & Appendix~\ref{sec:app-c} \\
  $\omega$, $\omega_0$, $\omega_{\min}$, $\gamma$ & the tree's width per round, start width, minimum width; width decay & Algorithm~\ref{alg:c3} \\
  \ckz $m$, $g$ & pruning margin; grace (children a node gets before it can be pruned) & Algorithm~\ref{alg:c3} \\
  \bottomrule
\end{tabularx}\par}

\section{Measurement and Calibration}
\label{sec:app-a}

\paragraph{Timing protocol.} Every candidate is timed with the benchmark's own harness on our B200 pods: 10 warm-up and 50 timed iterations, the L2 cache flushed and the inputs cloned before each iteration, in an isolated process. A candidate is correct only if it passes on every workload, and a partial run counts as a failure. Each timed run holds its GPU alone under a per-GPU lock (Appendix~\ref{sec:app-h}).

\paragraph{Clocks.} Our pods run the SM clock unlocked (\S\ref{sec:reward}), so local timings read fast and local SOL reads high. We call the site's latency divided by ours, for the same kernel, the \emph{host factor}. Over all 235 final kernels (as of 2026-11-10), the pod-timed geomean speedup over the official baseline is $5.30\times$, against the site-basis $4.47\times$ of \S\ref{sec:distribution} (there, L1, L2 and Quantization are site-timed and FlashInfer-Bench is timed locally). The host factor varies by kernel, so a single factor is only a first approximation.

\paragraph{Calibration is for people only.} People use a calibrated estimate of the site score to choose what to send. The agents never see the host factor, the estimate or any official score; the search optimizes raw local measurements. Each calibration rule was written down before we read the site results it was tested on, and was adopted only if it passed.

\paragraph{The score.} For each workload $w$ the site computes the score $S_w(k)$ of \S\ref{sec:challenge}. Judging from its results, the site caps each workload at 1, and a board's mean counts an unsolved problem as 0. Our implementation matched a real site result to four decimals (0.99715 against 0.997192).

\paragraph{The reward and aggregation.} Before the final run and for most of it, the agents maximized the official-baseline speedup $R(k)$ of \S\ref{sec:reward}. The SOL score was shown only as context and ranked nothing. The site-style score that ranked candidates late in the run is the arithmetic mean over workloads of $\min(1, S_w(k))$, computed from our own $t_k$ and the published $t_b$ and $t_{\mathrm{sol}}$. No official score of our kernels enters either version. Speedups in this report are geometric means over problems; SOL scores are arithmetic means; we report medians where a distribution matters and do not report per-class correction factors.

\section{Verification Checks}
\label{sec:app-b}

\paragraph{Official correctness on every workload.} Every evaluation runs the benchmark's own correctness test on every workload, with the official tolerance. A kernel that fails one workload, or a partial run, does not join the archive.

\paragraph{Robustness.} Passing the shipped workloads is not enough, so a robustness check reruns the kernel on shapes outside the shipped set (recombined sizes, sizes the benchmark never ships, and sizes above the shipped range) and on stress inputs with extreme or structured values. Only a ROBUST verdict passes, a timeout is not robust, and a failure of the reference code never counts against our kernel. During the run, plain code, not a model, re-checks each run's new best automatically, and large gains also get a memory-safety check. A best that fails stops counting as the run's best and the controller is told; its measured score is never changed. At send time the check is repeated with many random seeds on the exact bytes sent.

\emph{Coverage (as of \freezedate{}).} After the change to automatic checks (early in the run), robustness checks per landed candidate fell from 0.66 to 0.27 (\S\ref{sec:lesson-3}). Then 2.6\% of candidates got a robustness check, about 1 in 40, yet 94.9\% of the run's new bests were checked, usually within minutes of landing. Of the checked new bests, 95.9\% came back ROBUST and 1.8\% NOT\_ROBUST. A kernel that never becomes a run's best is not robustness-checked during the run; it still passes the official correctness test on every workload.

\paragraph{Precision.} The benchmark paper allows lower precision when data types match and tolerances are met~\citep{lin2026solexecbench}; during the run we adopted a stricter rule: (1) reduced-precision math counts only if the kernel keeps the declared input and output data types and passes the robustness check; (2) emulating fp32 on faster units (3xTF32, BF16x9, Ozaki schemes) is fine, while bf16 or fp16 math below the declared type is a violation; (3) Quantization problems are exempt, because their data types are the point of the problem.

\begin{table}[!htbp]
  \centering\cktablesetup\renewcommand{\arraystretch}{1}
  \caption{\textbf{The 12 checks before a send.} Every row we send must pass all of them. Checks 1 to 10 are ours and run on each row before it is sent; check 11 is a daily audit; check 12 is the site's own score. Some checks were added during the run (for example, check 2b).}
  \label{tab:send-checks}
  \begin{tabularx}{\linewidth}{@{\hspace{4pt}}rQ{2.8cm}W{1}@{\hspace{4pt}}}
    \toprule
    \thead{\#} & \thead{Check} & \thead{What it does} \\
    \midrule
    1 & Correctness & Benchmark harness on a B200, every workload, official tolerance; partial run fails \\
    \ckz 2 & Robustness & Extra shapes and special input values (above). Only ROBUST passes. \\
    2b & Multi-seed re-check & Robustness check, 16 seeds, exact bytes sent, two runs \\
    \ckz 3 & Memory safety & Memory and race checks (compute-sanitizer) on one small workload. \\
    4 & Precision & Declared dtypes kept; reduced precision only if ROBUST; Quantization exempt \\
    \ckz 5 & Reward hacks & Speed and correctness come from real work on the inputs \\
    6 & Output buffers & Output checks: outputs come from real work on every call. \\
    \ckz 7 & Build time & The kernel builds within the site's build-time limit, with a margin. \\
    8 & Site rehearsal & Official evaluation rehearsed on our B200 just before the send \\
    \ckz 9 & One review & One checklist review per row, with a written verdict \\
    10 & Send gates & The send tool refuses any row that is not ROBUST and clean. \\
    \ckz 11 & Agent audit & A daily audit of the agents' shell commands. \\
    12 & The site & The official site scores the kernel. Only that score counts. \\
    \bottomrule
  \end{tabularx}
\end{table}

\paragraph{Reward hacks.} The checks against reward hacks (\S\ref{sec:verify}; check 5 in Table~\ref{tab:send-checks}) are enforced in code, not only stated to the agents.

\paragraph{A fault of our own scorer.} For a period our scorer counted an evaluation that ran fewer workloads than it should as correct. When we found it, the affected rows were demoted and some problem bests went down.

\paragraph{Formal verification was ruled out.} In a research pass before the run, no formal or SMT-based tool we found modeled Blackwell's asynchronous tensor-core operations; the best candidate covered about 16\% of the problems. We chose testing outside the model and state the limitation in \S\ref{sec:conclusion}. 

\section{Search Algorithms}
\label{sec:app-c}\label{sec:app-c1}

\begin{figure}[!htbp]
  \centering
  \includegraphics[width=\linewidth,height=0.85\textheight,keepaspectratio]{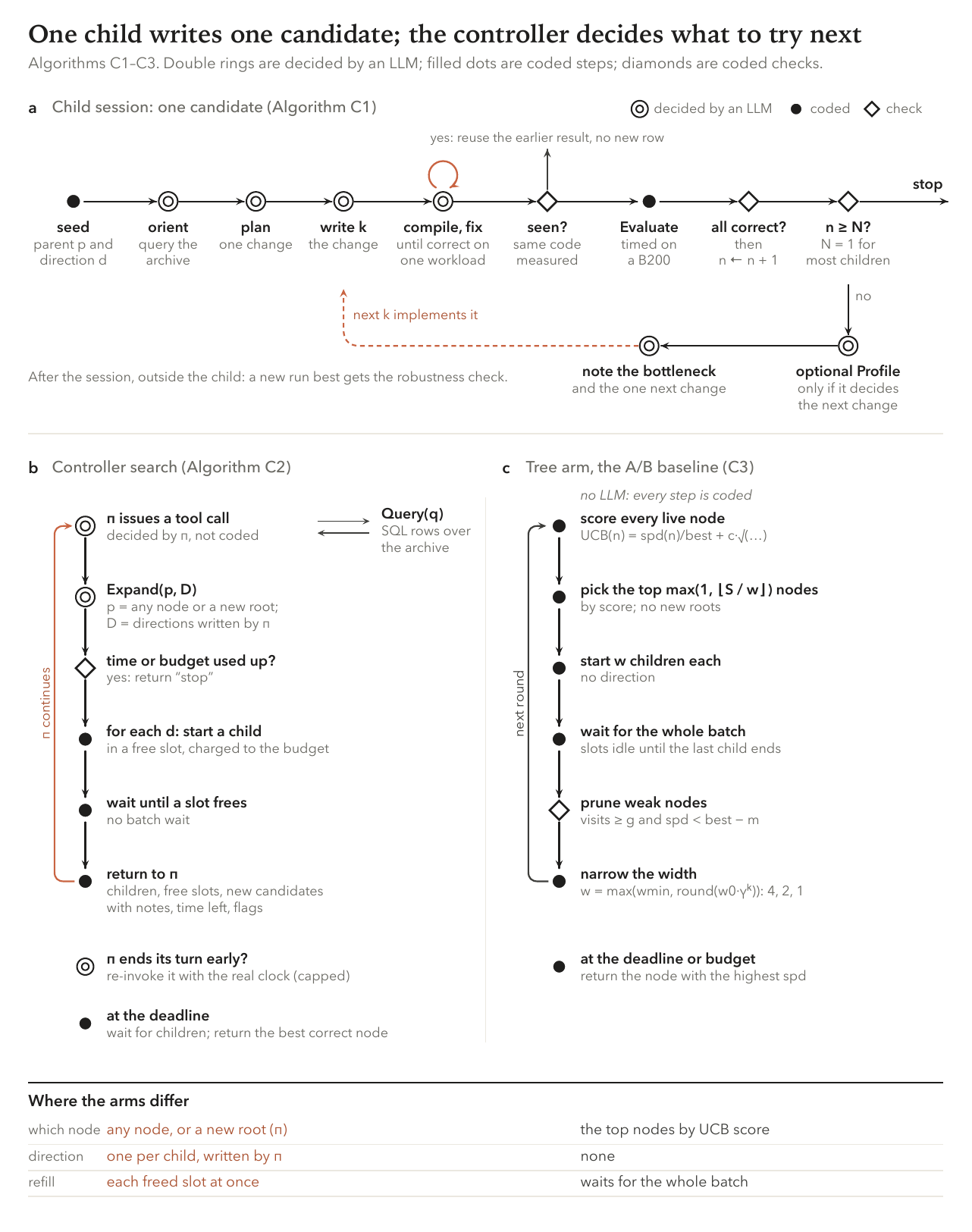}
  \caption{A child writes one candidate; the controller decides what to try next, and the tree arm replaces that choice with a coded score. Double rings are LLM decisions, filled dots coded steps, diamonds coded checks.}
  \label{fig:algo-flow}
\end{figure}

Figure~\ref{fig:algo-flow} gives the overview; Algorithms~\ref{alg:c1}--\ref{alg:c3} give the details.

The \hyperref[sec:notation]{Notation table} lists every symbol. Late in the run, $\mathrm{spd}(n)$ was the locally computed site-style score (Appendix~\ref{sec:app-a}). In the final run a child's default cap $N$ is one candidate, so a child normally stops at its first correct kernel and the controller makes most ``what next'' choices, one direction per child.

\FloatBarrier

\begin{algorithm}[!htbp]
\caption{Child session (one candidate).}
\label{alg:c1}
\small
\begin{algorithmic}[1]
\Require problem $P$, run $r$, parent $p$ or $\bot$, direction $d$ or $\bot$, cap $N$ \Comment{$N=1$ for most children}
\Require archive $A$ of run $r$; B200 tools \textsc{Compile}, \textsc{Evaluate}, \textsc{Profile}
\Ensure at most $N$ new correct nodes in $A$, each with an edge from its parent
\If{$p \neq \bot$} $\mathit{seed} \gets$ code and results of $p$ \Comment{never re-scored unchanged}
\EndIf
\State task: make one change to $\mathit{seed}$; with $d$, the change is $d$ (for a new root, $d$ is the approach)
\State orient: query this run's archive \Comment{best so far, dead ends, weak shapes}
\State plan the change
\State $\mathit{done} \gets 0$ \Comment{correct candidates so far}
\Loop
  \State write candidate $k$ (one change)
  \State \textbf{repeat} \textsc{Compile}$(k)$; fix $k$ \textbf{until} $k$ compiles and is correct on one workload
  \If{$k$ matches code already measured in this run} return the earlier result; no new row
  \EndIf
  \State $\mathit{res} \gets \textsc{Evaluate}(k)$ \Comment{full timed run on a B200; stores $k$ with its parent}
  \If{$\mathit{res}$ is correct on every workload} $\mathit{done} \gets \mathit{done}+1$
  \EndIf
  \If{$\mathit{done} \geq N$} \textbf{break} \Comment{the score never stops a child}
  \EndIf
  \State optional: \textsc{Profile}$(k)$, only if it decides the next change
  \State write down the bottleneck and the one next change; the next $k$ implements it
\EndLoop
\State after the session, outside the child: a new run best gets the robustness check (Appendix~\ref{sec:app-b})
\end{algorithmic}
\end{algorithm}

\begin{algorithm}[!htbp]
\caption{Controller search for one problem (final run).}
\label{alg:c2}
\small
\begin{algorithmic}[1]
\Require problem $P$, run $r$, child slots $\mathit{slots}$, deadline $T$, backstop budget $B$, LLM controller $\pi$ (the $\pi_{\mathrm{LLM}}$ of \S\ref{sec:archive})
\Require archive $A$ of run $r$ \Comment{after a warm restart it already holds this run's work}
\Ensure every child that started runs to its end; the best node is reported
\State $\mathit{used} \gets 0$
\State $\pi$.start(definition of $P$)
\While{$\pi$ issues tool calls} \Comment{the policy is $\pi$'s, not coded}
  \If{call is \textsc{Query}$(q)$} return the rows of SQL query $q$ over $A$
  \ElsIf{call is \textsc{Expand}$(p, D)$} \Comment{$p$: a parent node or a new root; $D$: directions}
    \If{now $\geq T$ \textbf{or} $\mathit{used} \geq B$} return ``stop''
    \EndIf
    \For{$d \in D$}
      \If{a slot is free \textbf{and} $\mathit{used} < B$}
        \State $\mathit{used} \gets \mathit{used}+1$; start \textsc{Child}$(P, r, p, d, N)$ in the background \Comment{Alg.~\ref{alg:c1}}
      \Else\ mark $d$ not started (not charged)
      \EndIf
    \EndFor
    \State wait until a slot frees (up to 15 min if $D$ is empty and a slot is free)
    \State return finished and running children, free slots, new candidates with their notes, minutes left, flags on bests that failed a check
  \EndIf
  \If{$\pi$ ends its turn with $> 10$ min and some budget left}
    \State re-invoke $\pi$ with the real clock \Comment{up to 20 counted times; API errors retried, not counted}
  \EndIf
\EndWhile
\State wait for all running children
\State \Return the correct node in $A$ with the highest reward
\end{algorithmic}
\end{algorithm}

The coded alternative in the controller-versus-tree A/B (\S\ref{sec:lesson-5}, Appendix~\ref{sec:app-e1}) is a fixed program with no LLM that picks which node to build on. It uses the same Claude child agents, without directions, and the same archive design. Each round it scores every live node with
\[\mathrm{UCB}(n) = \frac{\mathrm{spd}(n)}{\mathit{best}} + c\,\sqrt{\frac{\ln(|A|+1)}{\mathrm{visits}(n)+1}}, \qquad c = 0.7,\]
where $\mathrm{spd}(n)$ is node $n$'s measured speedup over $t_b$, $\mathit{best}$ is the top speedup among live nodes, $\mathrm{visits}(n)$ is the number of children node $n$ already has, and $|A|$ is the number of nodes in the archive. The score has the shape of an upper-confidence bound but is a priority rule, not a bandit estimate: each node is one fixed kernel with one measured speedup, and nothing is passed up the tree.

\begin{algorithm}[!htbp]
\caption{Tree arm with a UCB-style score (the A/B baseline).}
\label{alg:c3}
\small
\begin{algorithmic}[1]
\Require problem $P$, run $r$, start width $\omega_0$, decay $\gamma$, minimum width $\omega_{\min}$, child slots $\mathit{slots}$, margin $m$, grace $g$, $c = 0.7$, backstop $B$, deadline $T$, archive $A$ of run $r$
\If{$A$ has no node} try up to 3 cold roots \textsc{Child}$(P, r, \bot, \bot, 1)$
\EndIf
\State $i \gets 0$; $\mathit{pruned} \gets \emptyset$; $e \gets 0$ \Comment{$i$: round; $e$: children started}
\While{$e < B$ \textbf{and} now $< T$}
  \State $\omega \gets \max(\omega_{\min}, \mathrm{round}(\omega_0 \cdot \gamma^{i}))$ \Comment{this round's width}
  \State $\mathit{best} \gets$ highest $\mathrm{spd}$ over live nodes; $\mathrm{visits}(n) \gets$ number of nodes whose parent is $n$
  \State score every live node with $\mathrm{UCB}(n)$
  \State $\mathit{picks} \gets$ the top $\max(1, \lfloor \mathit{slots} / \omega \rfloor)$ live nodes by $\mathrm{UCB}$
  \For{each pick $n$}
    \State start $\omega$ children \textsc{Child}$(P, r, n, \bot, 1)$; $e \gets e + \omega$ \Comment{no direction}
  \EndFor
  \State wait for the whole batch
  \For{each $n \notin \mathit{pruned}$ that is not the best}
    \If{$\mathrm{visits}(n) \geq g$ \textbf{and} $\mathrm{spd}(n) < \max \mathrm{spd} - m$} prune $n$
    \EndIf
  \EndFor
  \State $i \gets i + 1$
\EndWhile
\State \Return the node with the highest $\mathrm{spd}$
\end{algorithmic}
\end{algorithm}

A/B settings: $\omega_0 = 4$, $\gamma = 0.6$, $\omega_{\min} = 1$ (widths 4, 2, 1), $g = 3$, $m = 0.15$ in speedup units.

\begin{figure}[!htbp]
  \centering
  \includegraphics[width=\linewidth]{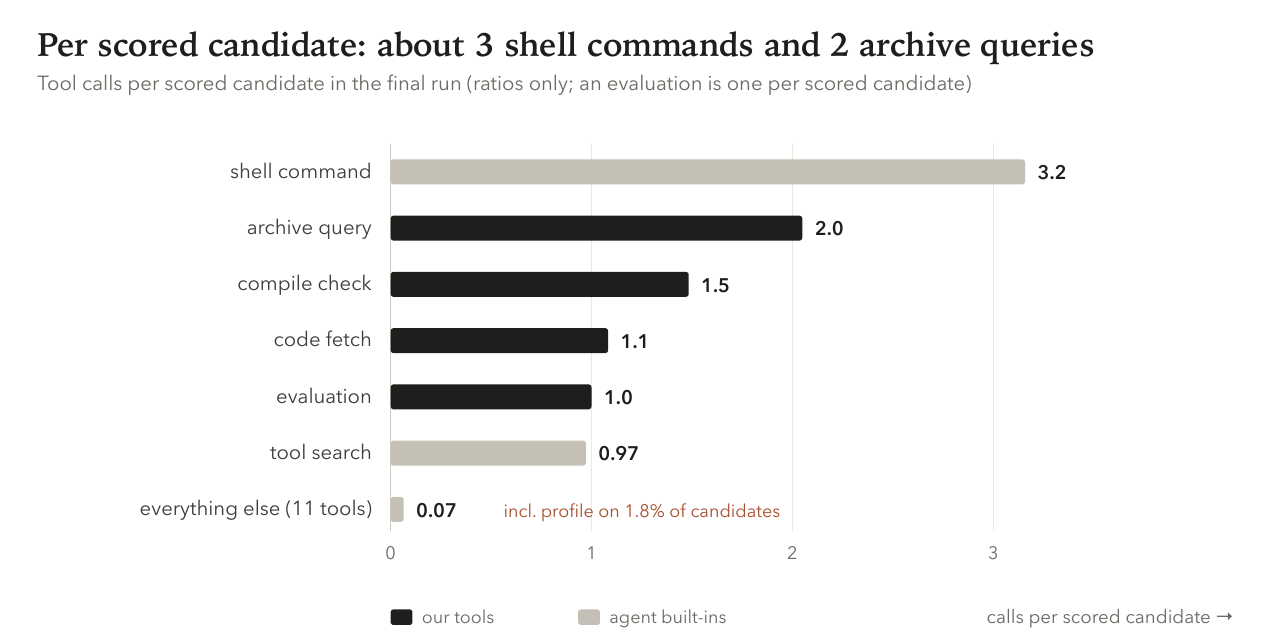}
  \caption{Per scored candidate, a child runs about 3 shell commands, 2 archive queries and 1.5 compile checks, and profiles on under 2\% of candidates. Ratios to the one evaluation per scored candidate.}
  \label{fig:tool-share}
\end{figure}

\paragraph{Agent instructions.} The controller and the children follow fixed instructions in our code, which we do not publish.

\section{Archive Schema}
\label{sec:app-d}

The archive is the search's memory: every scored kernel, every workload time, every profile, every robustness check and every agent step. Each run of each problem writes an append-only ledger plus one folder per candidate on shared storage. A candidate's files are written first and its ledger row last, so a row always points to complete files. The candidate id is a hash of the code, so a replay after a crash adds no duplicate. The database is a copy rebuilt from the ledgers, never written by agents.

\paragraph{Design choices.} (1) \emph{A copy, not a store}: the ledgers are the only place anyone writes, and the database is rebuilt from them (\S\ref{sec:archive}). (2) \emph{Agents only read}: the connection opens read-only and a SQL authorizer also refuses writes; we added the authorizer after a review found a write statement that slipped past an older check. (3) \emph{Each agent sees its own run}: an agent's query is answered from a small database built from its own run's ledger, so it is always current and holds nothing from other runs. (4) \emph{Big files stay outside}: code and profiles stay on shared storage, and the database points to each. (5) \emph{One schema for people and agents}: the same tables are exported for analytics, so the same SQL runs on both.

\begin{table}[!htbp]
  \centering\cktabledense\renewcommand{\arraystretch}{1}
  \caption{\textbf{The tables.} Most rows hang off a candidate (Figure~\ref{fig:F2b}b).}
  \label{tab:schema}
  \begin{tabularx}{\linewidth}{@{\hspace{4pt}}Q{2.7cm}W{1}W{1}@{\hspace{4pt}}}
    \toprule
    \thead{Table} & \thead{One row is} & \thead{Relations} \\
    \midrule
    \texttt{candidates} & one scored kernel in one run (a tree node) & points to its parent candidate in the same run (the tree edge; none for a root) \\
    \ckz \texttt{evaluation} & one candidate on one workload & belongs to a candidate; joins the published workload times \\
    \texttt{official\_anchors} & one workload of one problem, as published by the benchmark & holds the published baseline and SOL times \\
    \ckz \texttt{problem\_meta} & one problem & describes the problem behind every run \\
    \texttt{profiles} & one profiled candidate & belongs to a candidate \\
    \ckz \texttt{robustness} & one robustness-check result & belongs to a candidate \\
    \texttt{trace} & one step of one agent session & may name the candidate the step produced \\
    \bottomrule
  \end{tabularx}
\end{table}

\paragraph{Views.} Derived views give one row per run and problem, each run's best correct candidate, and the best correct candidate per problem across runs (people use this one; agents never did in the final run). The controller's archive tool also gives each node's descendant statistics, so the controller can judge a direction by its subtree.

\paragraph{What a candidate folder holds.} The kernel exactly as sent to the GPU tools, its evaluation result and its metadata. Agent sessions are stored as one transcript per session.

\section{A/B Tests in Full}
\label{sec:app-e}

\begin{table}[!htbp]
  \centering\cktabledense\renewcommand{\arraystretch}{1}
  \caption{\textbf{The design tests behind \S\S\ref{sec:lessons}--\ref{sec:design}.} One row per test; details follow in E.1--E.6. Arms ran at the same time on the same fleet unless noted. Phase 1 and Phase 2 are pre-final harness versions (Appendix~\ref{sec:app-j}).}
  \label{tab:ab-full}
  \begin{tabularx}{\linewidth}{@{\hspace{4pt}}W{1.3}lW{0.8}W{1.1}W{0.8}@{\hspace{4pt}}}
    \toprule
    \thead{Test} & \thead{Problems (n)} & \thead{Horizon} & \thead{Result} & \thead{Verdict} \\
    \midrule
    E.1 Controller vs coded tree & 38 paired & 24 h per arm & controller better on 35 of 38 at equal time; at equal count, better on 15, worse on 23 & Controller adopted \\
    \ckz E.2 Which engine & 4; then 38 pairs & 600 min; E.1 & engines within $\sim$3\% & Claude inside controller \\
    E.3 Mandatory references & 233 paired & $\sim$23 h of 48 h & ratio 0.981; p-value $\approx 0.24$ & No measured lift \\
    \ckz E.4 Structural steering & 4 stuck runs & 210 min & $-$0.30\% to +0.05\% & Screening null; not shipped \\
    E.5 Share sibling designs (FI) & 24 paired & about a day & sharing better on 9 of 24 & No gain; left off \\
    \ckz E.6 Design-switch mode (FI) & 4 vs 4 & short window & 77.8\% vs 3.6\% switched & Kept on FI \\
    \bottomrule
  \end{tabularx}
\end{table}

\subsection{LLM Controller vs Coded Tree}
\label{sec:app-e1}

\emph{Question.} Does LLM node selection beat a coded UCB-style tree (Algorithm~\ref{alg:c3}) with the same Claude child agents and the same archive design?
\emph{Setup.} Phase 1 harness; 40 problems stratified over all four boards (L1 16, L2 14, Quantization 6, FlashInfer-Bench 4), 38 correct in both arms; one run per arm with a 24 h budget (\S\ref{sec:lesson-5}). Run spans (first to last ledger row) had a median of 20.5--21.5 h in both arms, so ``24 h'' is the budget, not the measured span.
\emph{Measure and result.} Best official-baseline speedup per problem, paired. \S\ref{sec:lesson-5} gives the win counts at equal time and at equal candidate count. At equal wall-clock time the geomean ratio was 1.143 (sign test p-value $\approx 7\times10^{-8}$).
\emph{Why it wins.} The lead comes from the attempt rate (\S\ref{sec:lesson-5}). The arms differ in batch width, directions and pruning: the tree narrows its width quickly (4, 2, 1), leaving slots idle, while the controller keeps dispatching full batches. In round 2 (4 problems, 600 min per configuration), tree and controller tied 2--2, with margins of 1--3\% on three of the four.

\subsection{Which Engine Writes Kernels}
\label{sec:app-e2}

\emph{Question.} Which model should the children run?
\emph{Setup.} Rounds 1--2 on 4 problems (one per board), then the 38 pairs of E.1.
\emph{Measure and result.} Best speedup. In round 2 (600 min each) the three engines' best speedups ended within about 3\%; in round 1 Codex made 4--7$\times$ more candidates and did not win. In E.1, inside the controller, Claude beat the mix 28--10 and Codex 31--7; inside the tree, the mix beat Claude 25--13.
\emph{Confounds and scope.} Engine and steering interact, and round 1 budgets were unequal. We kept Claude inside the controller.

\subsection{Mandatory References}
\label{sec:app-e3}

\emph{Question.} Does making the optional references mandatory raise final speedups?
\emph{Setup.} Phase 2 harness, run as in \S\ref{sec:design-guidance}; 233 problems paired; the test stopped after about 23 h of a planned 48 h.
\emph{Measure and result.} Ratio of best speedups, gated over control (Table~\ref{tab:design}): geomean 3.87 vs 3.94; sign test p-value $\approx 0.24$. The arms also crossed: on 36 problems the gated arm led early and the control arm passed it later, and the reverse on 34.
\emph{Confounds and scope.} The gated arm lost some host time for about half the test to a watchdog bug and was throttled harder on model calls (12.3\% vs 8.2\% of children ended by throttling); both confounds cut against it, so a small benefit is not ruled out. References stayed optional; \S\ref{sec:design-guidance} gives their use in the final run.

\subsection{Structural Steering of Stuck Runs}
\label{sec:app-e4}

\emph{Question.} Does an added instruction to try a structural change help runs that have plateaued?
\emph{Setup.} A final-run snapshot; 4 stuck runs, on and off arms, one replicate each, a fresh relaunch in both arms; 210 min.
\emph{Measure and result.} Pre-registered score, on minus off: $-0.23\%$, $+0.05\%$, $-0.03\%$ and $-0.30\%$.
\emph{Confounds and scope.} The manipulation check failed: almost no structural changes happened in either arm, so the test is a screening null.

\subsection{Sharing Sibling Designs on FlashInfer-Bench}
\label{sec:app-e7}

\emph{Question.} Does showing a run the designs of its sibling runs help?
\emph{Setup.} The final run; 26 FlashInfer-Bench problems, 24 with a robust best in both arms; about a day; one replicate.
\emph{Measure and result.} Robust-best local SOL, paired: sharing better on 9, control on 15.
\emph{Confounds and scope.} One replicate. No gain; left off.

\subsection{Stuck Detection and a Design-switch Mode on FlashInfer-Bench}
\label{sec:app-e8}

\emph{Question.} Does a design-switch mode make stuck runs try another design family?
\emph{Setup.} Late in the final run; 4 vs 4 matched runs; a short window.
\emph{Measure and result.} 77.8\% vs 3.6\% of directed children switched design family; the score effect was inconclusive.
\emph{Confounds and scope.} Short window and small n. Kept on FlashInfer-Bench and judged by its score trend.

\section{More Results}
\label{sec:app-f}

\begin{figure}[!htbp]
  \centering
  \includegraphics[width=\linewidth]{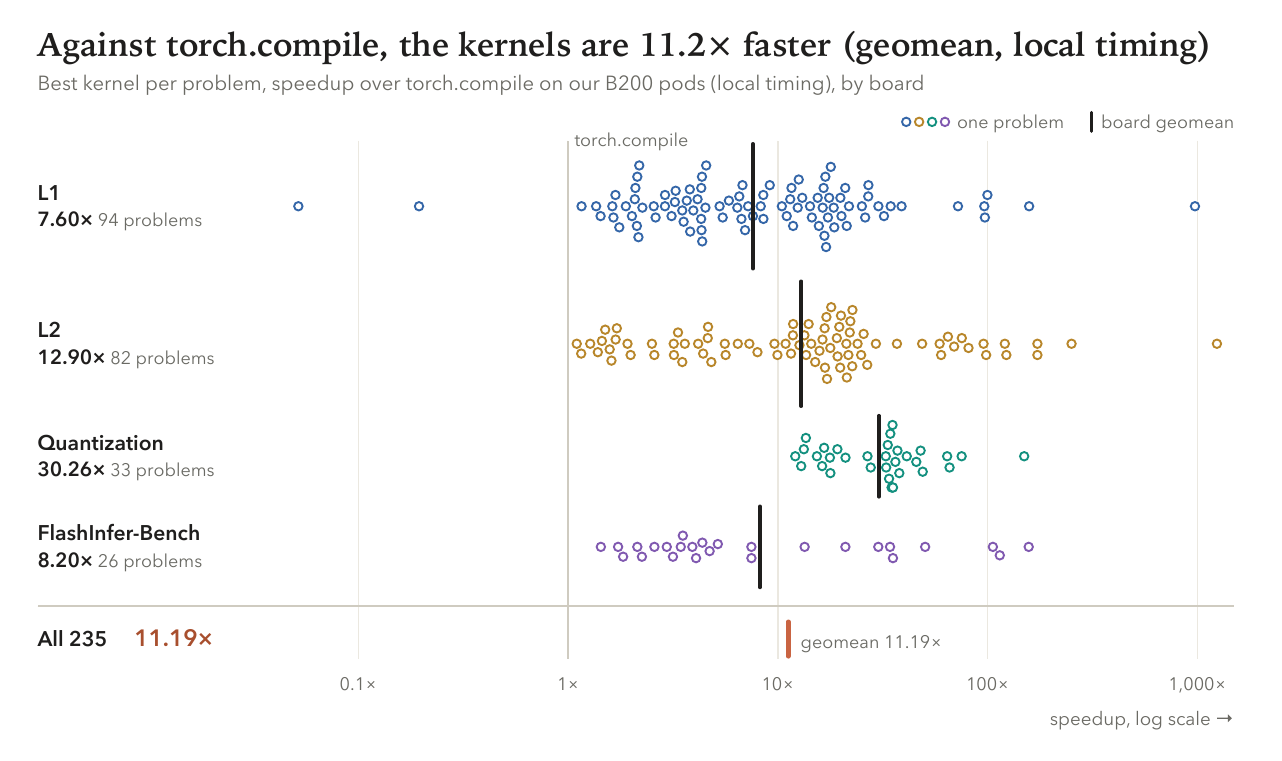}
  \caption{Against torch.compile, our kernels are 11.2$\times$ faster (geomean over 235 problems). Local timing on our B200 pods, kernels as of 2026-11-10; each tick is a board's geomean.}
  \label{fig:F3b}
\end{figure}

\begin{figure}[!htbp]
  \centering
  \includegraphics[width=\linewidth]{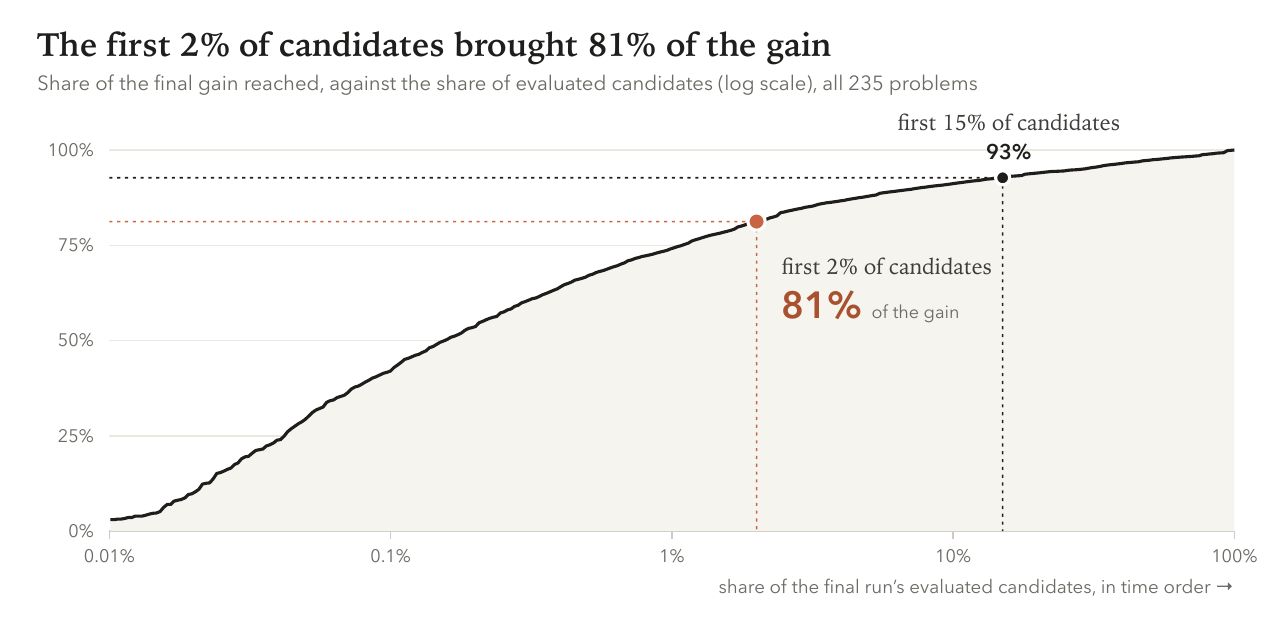}
  \caption{Most of the gain also came early in candidates: the first 2\% brought 81\% of it, the first 15\% brought 93\%. Gain is the best local SOL so far minus the first correct kernel's, pooled; log x.}
  \label{fig:F4b}
\end{figure}

\paragraph{Against \texttt{torch.compile}.} Figure~\ref{fig:F3b} gives each board's speedup over \texttt{torch.compile} (default mode); \S\ref{sec:distribution} explains why it differs from the speedup over the benchmark's baseline. Figure~\ref{fig:F4b} gives the climb against evaluated candidates.

The tables below give the site's scores of our best kernel per problem, including kernels not on the public board. They are therefore not the public board scores of \S\ref{sec:results}. The tables use kernels as of 2026-10-31. The counts in the next paragraph use our public rows on the boards as of \boarddate{}. The per-problem \#1s under Languages use kernels and boards as of 2026-11-10.

\paragraph{SOL follows the speedup.} Across the 209 problems of the three boards we submitted to, SOL and speedup over the baseline rank the problems in almost the same order (Spearman 0.91), since a kernel exactly as fast as the baseline scores 0.5 on a workload. \S\ref{sec:kernels} gives the problems below $1.5\times$.

\begin{table}[!htbp]
  \centering\cktablesetup\renewcommand{\arraystretch}{1.25}
  \caption{\textbf{Quantization is closest to the speed-of-light bound; L1 and L2 have longer low tails.} SOL of our best kernel per problem, by board.}
  \label{tab:sol-by-board}
  \begin{tabular}{@{\hspace{4pt}}lrrrcr@{\hspace{4pt}}}
    \toprule
    \thead{Board} & \thead{Problems} & \thead{Mean} & \thead{Median} & \thead{p25--p75} & \thead{$\ge 0.9$} \\
    \midrule
    L1 & 94 & 0.815 & 0.844 & 0.743--0.915 & 31\% \\
    \ckz L2 & 82 & 0.817 & 0.831 & 0.719--0.930 & 38\% \\
    Quantization & 33 & 0.909 & 0.920 & 0.866--0.967 & 64\% \\
    \ckz All three & 209 & 0.830 & 0.856 & 0.756--0.929 & 39\% \\
    \bottomrule
  \end{tabular}
\end{table}

\begin{table}[!htbp]
  \centering\cktablesetup\renewcommand{\arraystretch}{1.25}
  \caption{\textbf{Plain GEMM work and full layers are furthest from SOL; attention holds the largest share of the total gap.} Problems grouped by keywords in their names (first match in footnote order).}
  \label{tab:sol-by-type}
  \begin{tabular}{@{\hspace{4pt}}lrrrrr@{\hspace{4pt}}}
    \toprule
    \thead{Type$^a$} & \thead{Problems} & \thead{Mean SOL} & \thead{Median SOL} & \thead{SOL $<0.8$} & \thead{Share of gap} \\
    \midrule
    GEMM, projection, MLP & 19 & 0.766 & 0.767 & 13 & 13\% \\
    \ckz Full layer or block & 27 & 0.772 & 0.746 & 15 & 17\% \\
    Attention & 63 & 0.810 & 0.841 & 24 & 34\% \\
    \ckz Normalization & 8 & 0.818 & 0.844 & 3 & 4\% \\
    Elementwise and other & 11 & 0.844 & 0.907 & 4 & 5\% \\
    \ckz RoPE, position, embedding & 22 & 0.865 & 0.909 & 6 & 8\% \\
    Convolution & 6 & 0.871 & 0.859 & 0 & 2\% \\
    \ckz Scan, SSM, recurrence & 17 & 0.886 & 0.902 & 3 & 5\% \\
    MoE and routing & 36 & 0.887 & 0.911 & 7 & 11\% \\
    \bottomrule
  \end{tabular}
  \cknotes{$^a$ First match on the problem name, in this order: scan, SSM or recurrence (mamba, ssm, scan, hyena, time decay, delta rule, segsum); MoE and routing (moe, expert, routing, topk, dispatch); full layer or block (layer, block); attention; convolution (conv, unet, vae, resnet, pyramid); RoPE, position or embedding; normalization (norm); GEMM, projection or MLP (projection, matmul, linear, mlp, ffn, feedforward, gate up, swiglu, geglu, gemm); the rest are elementwise and other. ``Share of gap'' is the type's part of $\sum(1-\mathrm{SOL})$ over the 209 problems.}
\end{table}

\paragraph{Forward and backward.} In L2, the 21 backward problems score lower than the 61 forward ones (mean SOL 0.77 against 0.83). In L1 it is the other way round: 0.85 for the 25 backward problems and 0.80 for the 69 forward ones.

\begin{table}[!htbp]
  \centering\cktablesetup\renewcommand{\arraystretch}{1}
  \caption{\textbf{The agents wrote Blackwell-native code.} Share of the 235 best kernels whose source contains each pattern (a text match; large files keep old and shape-specific paths, so a pattern in the source is not proof that it runs on the scored path). Kernels as of 2026-10-31.}
  \label{tab:techniques}
  \begin{tabularx}{\linewidth}{@{\hspace{4pt}}W{1}rrrrr@{\hspace{4pt}}}
    \toprule
    \thead{Technique (pattern in source)} & \thead{L1} & \thead{L2} & \thead{Quant.} & \thead{FlashInfer} & \thead{All 235} \\
    \midrule
    \texttt{tcgen05} instructions named & 60\% & 98\% & 94\% & 54\% & 77\% \\
    \ckz \texttt{tcgen05.mma} in code & 52\% & 94\% & 88\% & 46\% & 71\% \\
    TMEM allocated & 54\% & 95\% & 94\% & 46\% & 73\% \\
    \ckz TMA tensor copy & 53\% & 94\% & 88\% & 54\% & 72\% \\
    2-SM MMA (\texttt{cta\_group::2}) & 47\% & 80\% & 73\% & 31\% & 60\% \\
    \ckz Thread-block clusters & 68\% & 88\% & 88\% & 85\% & 80\% \\
    \texttt{mbarrier} & 76\% & 95\% & 97\% & 81\% & 86\% \\
    \ckz Warp-specialized register split (\texttt{setmaxnreg}) & 17\% & 30\% & 58\% & 19\% & 28\% \\
    Older \texttt{mma.sync} path & 27\% & 52\% & 6\% & 69\% & 37\% \\
    \ckz Hopper \texttt{wgmma} & 0\% & 0\% & 0\% & 0\% & 0\% \\
    Inline PTX & 86\% & 99\% & 97\% & 100\% & 94\% \\
    \ckz Raw CUDA and PTX only (no CUTLASS or CuTe include, no library call) & 49\% & 38\% & 42\% & 58\% & 45\% \\
    PDL (programmatic dependent launch) & 86\% & 98\% & 97\% & 96\% & 93\% \\
    \ckz CUDA graph API & 50\% & 63\% & 64\% & 12\% & 52\% \\
    CUTLASS or CuTe C++ include & 24\% & 32\% & 45\% & 23\% & 30\% \\
    \ckz cuBLAS or cuBLASLt call & 15\% & 39\% & 3\% & 4\% & 20\% \\
    ATen or torch op call & 12\% & 22\% & 9\% & 12\% & 15\% \\
    \ckz FP8 types & 2\% & 6\% & 100\% & 4\% & 17\% \\
    FP4 / NVFP4 types & 0\% & 1\% & 45\% & 0\% & 7\% \\
    \ckz CuTe DSL & 11\% & 2\% & 6\% & 12\% & 7\% \\
    Triton & 7\% & 0\% & 3\% & 4\% & 4\% \\
    \bottomrule
  \end{tabularx}
\end{table}

\paragraph{Languages.} \S\ref{sec:kernels} gives the declared languages. Of the 78 per-problem \#1s of 2026-11-10, 74 are CUDA C++ and 4 are CuTe DSL. About 45\% of the kernels are raw CUDA and PTX (Table~\ref{tab:techniques}); about 31\% call a library somewhere, most often on the large L2 layer problems.

\section{Case-Study Details}
\label{sec:app-g}

Table~\ref{tab:case-milestones} lists the milestones of each case. \emph{Local} timings are from our B200 pods against the official baseline; \emph{site} values are the official site's. Times are shares of the final run's time (0\% = launch, 100\% = end of the data window); ``later'' means after that window.

{\cktabledense\renewcommand{\arraystretch}{1}
\begin{xltabular}{\linewidth}{@{\hspace{4pt}}llW{1}Q{2.9cm}r@{\hspace{4pt}}}
  \caption{Case milestones.}
  \label{tab:case-milestones}\\
  \toprule
  \thead{Case} & \thead{Basis} & \thead{Milestone} & \thead{Value} & \thead{Time (\% of run)} \\
  \midrule
  \endfirsthead
  \multicolumn{5}{@{}l}{\emph{Table~\thetable{} (continued)}}\\
  \toprule
  \thead{Case} & \thead{Basis} & \thead{Milestone} & \thead{Value} & \thead{Time (\% of run)} \\
  \midrule
  \endhead
  \bottomrule
  \endlastfoot
  Quant/030 & local & root & 6.254 & 0.7\% \\
  \ckz Quant/030 & local & parent before dip (path step 7) & 9.708 & 1.9\% \\
  Quant/030 & local & dip (path step 8; first CuTe tcgen05 attention) & 1.517 & 2.1\% \\
  \ckz Quant/030 & local & spill fix (path step 9) & 7.099 & 2.3\% \\
  Quant/030 & local & run best when step 10 landed (the run's incumbent) & 12.926 & 3.1\% \\
  \ckz Quant/030 & local & new attention rebased onto incumbent (path step 10) & 13.966 & 3.3\% \\
  Quant/030 & local & last node before timer cleanup (path step 155) & 25.364 & 34\% \\
  \ckz Quant/030 & local & timer cleanup node (path step 156) & 23.879 & 34\% \\
  Quant/030 & local & timer-free node later shipped (path step 161) & 25.419 & 36\% \\
  \ckz Quant/030 & local & best at the case-study cut (path step 270) & 27.357 & 96\% \\
  Quant/030 & local & best in a later snapshot (path step 295) & 27.570 & later \\
  \ckz Quant/030 & local & best of nodes NOT descended from the dip node & 24.874 & 31\% \\
  Quant/030 & share & nodes descended from the dip node & 93\% &  \\
  \ckz Quant/030 & site & shipped row SOL / site-timed speedup / workloads faster & 0.971073 / 20.67x / 16 of 16 & 36\% \\
  L2/001 & local & fastest node before dip (NOT\_ROBUST (failed some extra shapes)) & 5.733 & 4.7\% \\
  \ckz L2/001 & local & robustness fix (dip start) & 4.257 & 5.1\% \\
  L2/001 & local & valley min (ROBUST 6/6) & 4.170 & 5.6\% \\
  \ckz L2/001 & local & first node above pre-dip best & 5.737 & 8.3\% \\
  L2/001 & local & best & 15.198 & 96\% \\
  \ckz L2/001 & local & best of nodes NOT descended from the fix node & 13.512 &  \\
  L2/001 & share & nodes descended from the fix node & 86\% &  \\
  \ckz L2/001 & site & shipped row SOL / site-timed speedup / workloads faster & 0.936556 / 11.88x / 16 of 16 & 78\% \\
  L2/043 & local & run best when fresh root opened (multi-kernel chain) & 32.085 & 11\% \\
  \ckz L2/043 & local & fresh root (single fused persistent tensor-core kernel) & 15.096 & 11\% \\
  L2/043 & local & hybrid: incumbent chain + fused kernel on one shape class (path step 3) & 32.147 & 11\% \\
  \ckz L2/043 & local & best & 42.858 & later \\
  L2/043 & local & best of nodes NOT descended from the fresh root & 35.804 &  \\
  \ckz L2/043 & share & nodes descended from the fresh root & 97\% &  \\
  L2/043 & site & shipped row SOL / site-timed speedup / workloads faster & 0.971494 / 29.83x / 16 of 16 & 68\% \\
  \ckz L1/065 & local & run best (small-shape non-tensor-core kernel on a sibling branch) & 1.926 & 1.3\% \\
  L1/065 & local & tensor-core line node behind the leader & 1.397 & 1.5\% \\
  \ckz L1/065 & local & hybrid: leader kernel for small shapes + tensor-core otherwise (path step 3) & 2.097 & 1.8\% \\
  L1/065 & local & best & 5.626 & later \\
  \ckz L1/065 & local & best of nodes NOT descended from the behind node & 4.289 &  \\
  L1/065 & share & nodes descended from the behind node & 96\% &  \\
  \ckz L1/065 & site & shipped row SOL / site-timed speedup / workloads faster & 0.831881 / 4.28x / 16 of 16 & 54\% \\
  Quant/029 & local & best at about 1\% / 6\% / 100\% of the run, and later & 5.65 / 11.32 / 14.55 / 14.70 &  \\
  \ckz Quant/029 & site & first accepted send (unity build; still had runtime timing selection) & 0.926970 & 37\% \\
  Quant/029 & site & clean replacement (pinned + pruned to 4 translation units) & 0.931935 / 11.03x / 16 of 16 & 60\% \\
  \ckz Quant/029 & local & same bytes as the clean replacement on our pods (SOL / speedup vs official baseline) & 0.948 / 13.69x & 60\% \\
\end{xltabular}
}

\section{Infrastructure and Monitoring}
\label{sec:app-h}

This appendix describes the parts of the system by role, so the scaling lessons in \S\ref{sec:lessons} can be read against them. Names, sizes and identifiers of internal systems are left out.

\paragraph{Agent pods (CPU).} Controllers and children run as Claude sessions on CPU pods in a container cluster; the model is reached through a managed model API with pod-scoped credentials, so there are no long-lived keys. Model-call throttling limited early scale; spreading the load fixed it. A self-restarting job recreates a lost agent pod, and its runs resume from their own ledgers. The launcher tells a lost GPU pod from a fleet-wide outage: it recreates the first and waits out the second.

\paragraph{GPU evaluation pods.} Each B200 pod holds a whole GPU node and runs the benchmark's own harness plus our tools to compile, time and profile a kernel. Agents never run kernels themselves: each tool call is a remote command into a GPU pod of the agent's own run. Builds run first, outside the GPU lock, into a cache; the timed run then takes a per-GPU lock held only around the GPU phases. After the first warm restart, the mean hold per locked call fell 74\% (Figure~\ref{fig:F5} shows the lock time per scored candidate). The GPU pods hold no cloud credentials, the cloud metadata service is blocked in every pod, and agents may reach only the named GPU pods of their own run.

\paragraph{Shared storage and analytics.} Every run writes to a shared file system (Appendix~\ref{sec:app-d}). The final run used a fresh storage root with the benchmark inputs mounted read-only. One loop builds the shared database from all ledgers and exports it as columnar files to object storage, where a single scheduled writer merges them into tables that people query with SQL. A new kernel reaches these tables about ten minutes after its ledger row, plus build time.

\paragraph{Dashboard (for people only).} A web app reads the exported files read-only, has no cloud credentials and is reached only through a local port-forward. Appendix~\ref{sec:app-views} describes its six views, including the calibrated site estimate (Appendix~\ref{sec:app-a}), and shows five of them (Figures~\ref{fig:dash-progress}--\ref{fig:dash-sql}). The agents are not given the dashboard, nothing from it enters a prompt or a run folder, and the daily shell audit checks that no agent reached it.

\paragraph{Monitor.} A rule-based monitor runs on every snapshot. It has 23 stateless rules plus one recheck flag, each at one of three levels (info, warn, decision). Eleven \emph{operations} rules watch our infrastructure: a stale or failed export, a silent or slow host, a silent fleet or run, the share of children ended by model-call throttling, GPUs idle compared with their peers, launches and the run window. Twelve \emph{results} rules watch the search: the run behind an earlier run's curve, a problem far behind its own earlier run, no correct kernel after some hours, falling correctness fleet-wide or per run, correct rows without a score, a best that may not be valid, a long plateau (information only, never a stop signal), tool errors, the gap to the board's \#1, problems far from SOL, and a ranking of problems that spend their time badly. Thresholds were set from our own earlier runs (for example, the longest gaps of healthy hosts against real outages) and were tuned during the run, each time for a dated cause. No rule acts by itself: each issue goes to an owner, and slowness is never treated as a fault.

\paragraph{Alerts.} A notifier posts the monitor's findings to a team chat channel in three tiers: decision-level items at once, a warning summary at most every six hours and only if something changed, and a daily digest. It posts no reminders and no ``resolved'' lines. It posts a few lines a day. Because a quiet channel can also mean a dead notifier, a missing daily digest is itself the signal.

\begin{figure}[!htbp]
  \centering
  \includegraphics[width=\linewidth]{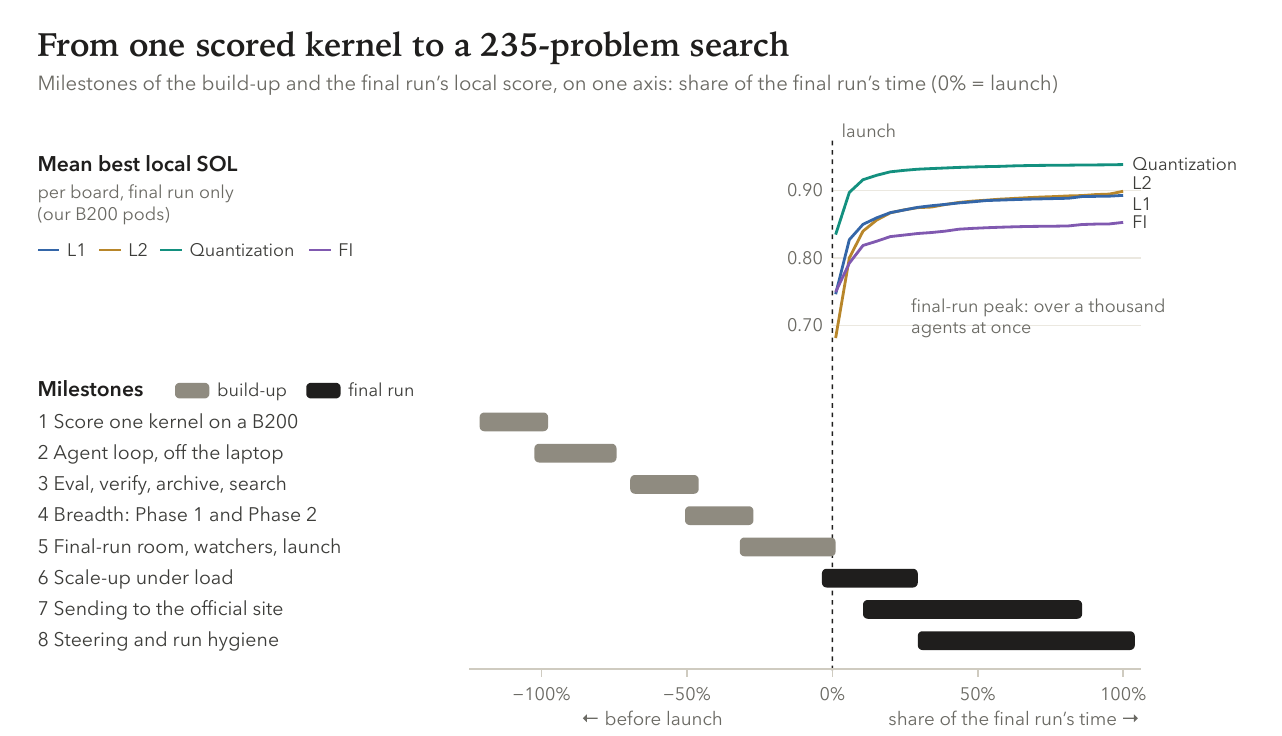}
  \caption{From one scored kernel to a 235-problem search: eight milestones (five before launch) and the final run's local score per board. The x-axis is the share of the final run's time (0\% = launch).}
  \label{fig:milestones}
\end{figure}

\paragraph{Hands-off by design.} Short-lived API errors are retried; a controller that stops early is woken again; a best that fails the robustness check is replaced by the next best. Warm restarts apply one change per launch and keep the run and its archive; a change to the agents alone keeps the GPU pods and cycles only the agent pods, which cut agent downtime from hours to minutes (Figure~\ref{fig:milestones}).

\section{Related Systems in Full}
\label{sec:app-i}

{\cktabledense\renewcommand{\arraystretch}{1}
\begin{xltabular}{\linewidth}{@{\hspace{4pt}}W{1.25}W{0.8}W{1.5}W{0.8}W{0.95}W{0.9}@{\hspace{4pt}}}
  \caption{\textbf{LLM and agent systems for GPU kernels (as of 2026-11-09).} ``-'' = not stated in the source.}
  \label{tab:related-full}\\
  \toprule
  \thead{System} & \thead{Trains model?} & \thead{Search structure} & \thead{Hardware} & \thead{Benchmark} & \thead{Anti-cheat measures} \\
  \midrule
  \endfirsthead
  \multicolumn{6}{@{}l}{\emph{Table~\thetable{} (continued)}}\\
  \toprule
  \thead{System} & \thead{Trains model?} & \thead{Search structure} & \thead{Hardware} & \thead{Benchmark} & \thead{Anti-cheat measures} \\
  \midrule
  \endhead
  \bottomrule
  \endlastfoot
  ForgeKernel~\citep{chen2026forgekernel} & No (fixed frontier models) & Architect$\to$coder milestones; linear cross-run generations; knowledge base & H100, B200, Ascend & Expert libraries; SOL-ExecBench FlashInfer-Bench; CANN-Bench & Tolerance tests; rejects reference delegation; state-decay inputs \\
  \ckz Cursor $\times$ NVIDIA~\citep{cursor2026multiagentkernels} & No (model not stated) & Planner + many workers, markdown protocol; reports no developer intervention in its debug/\allowbreak{}test/\allowbreak{}optimize loop; about 3 weeks, one run per language (CUDA C + inline PTX, CuTe DSL) & 27$\times$ B200 & SOL-ExecBench 235 & Benchmark's SOL-violation invalidation \\
  Recursive~\citep{recursive2026firststeps} & Not stated & Many long-horizon research threads, cross-task transfer & B200 & SOL-ExecBench 235 & Iterated reward-hack detector; strict audits; manual and AI review \\
  \ckz KernelArc~\citep{kundu2026kernelarc} & No & Strategy-specialized parallel agents, shared conclusions memory & H100, B200 & SOL-ExecBench (representative subset) & Deterministic benchmark guard \\
  DuoDiscover~\citep{agarwal2026duodiscover} & No & Agent fleet, proxy-evaluator hill-climbing, orchestrator seeds agents & GPU and TPU suites & SOL-ExecBench 235 and others & Fidelity tests re-check proxies against the true objective \\
  \ckz AVO~\citep{chen2026avo} & No & Agent as variation operator over a lineage, 7 days & B200 & MHA/\allowbreak{}GQA against cuDNN/\allowbreak{}FA4 & - \\
  KDA~\citep{nvlabs2026kda} & No & Agent workflow + KernelWiki & B200, B300 & FlashInfer contest & - \\
  \ckz JustGoal~\citep{chen2026justgoal} & No & Single goal-only agent & B200 & 10 operators against libraries & Reference viewing and copying prohibited \\
  ModelBlaster~\citep{dong2026modelblaster} & No weights (in-context knowledge base) & Actor-critic textual-gradient knowledge base & L40S, B200, Adreno, RVV & KernelBench; SOL-ExecBench L1+L2 & LLM soft-verification agent \\
  \ckz EAT~\citep{cao2026eat} & No & Supervisor reorganizes agent teams on plateau & H100, B200, MI350X & FlashInfer-Bench workloads & - \\
  AKO~\citep{luo2026ako} & No & Multi-agent; a human redirects stalled searches & B200 & 3 FlashInfer-Bench workloads & Human-enforced anti-hacking rules \\
  \ckz Harness Engineering~\citep{shui2026harness} & No & Human-written skills + Codex/\allowbreak{}Claude Code; bounded controller & B200 & FlashInfer contest, 5 definitions & Official-aligned timing harness \\
  K-Search~\citep{cao2026ksearch} & No & Co-evolving world model + evolution & H100, B200 & FlashInfer GQA/\allowbreak{}MLA/\allowbreak{}MoE; TriMul & - \\
  \ckz CAKE~\citep{ye2026cake} & No & Agents author a typed IR; the harness evolves & Ampere to B200 & Flash-KMeans, KDA, KNN & Verifier rules grown from failures \\
  KernelFalcon~\citep{wang2025kernelfalcon} & No & Deterministic orchestrator, parallel early-exit workers & NVIDIA (not stated) & KernelBench L1--3 (correctness) & Triton-only kernels; trusts LLM-written test harness \\
  \ckz CudaForge~\citep{zhang2025cudaforge} & No & Coder + judge loop with profiler feedback & A100/RTX & KernelBench & - \\
  AI CUDA Engineer / robust-kbench~\citep{lange2025robustkbench} & No & Translate, then evolutionary meta-generation + LLM verifiers & H100 & KernelBench $\to$ robust-kbench & Hardened benchmark after its speedups proved to be loophole exploits; LLM verifier \\
  \ckz METR KernelAgent~\citep{metr2025kernel} & No (plus one fine-tune ablation) & Parallel tree search, 300 attempts & H100 & Filtered KernelBench + level 5 & Task and solution filtering \\
  Kevin-32B~\citep{baronio2025kevin} & Yes (multi-turn RL) & 16 parallel trajectories $\times$ 8 refinement turns & H100/H200 & KernelBench L1--2 & Zero reward for PyTorch use or missing CUDA \\
  \ckz CUDA-L1~\citep{li2025cudal1} & Yes (contrastive RL) & - & A100 & KernelBench & Reward-hacking fixes found in training (timing sync, lazy-tensor checks) \\
  CUDA Agent~\citep{dai2026cudaagent} & Yes (agentic RL) & Agent in a skill-augmented environment & H20 & KernelBench & Automated verification and profiling \\
  \ckz TritonRL~\citep{woo2025tritonrl} & Yes & - & L40S & KernelBench & Multi-layer verification \\
  Kernel-Smith~\citep{du2026kernelsmith} & Yes (from evolution traces) & Evolutionary population & NVIDIA, MetaX & KernelBench-Triton & Backend-specific evaluation services \\
  \ckz TTT-Discover~\citep{yuksekgonul2026tttdiscover} & Yes (test-time RL) & Per-problem RL & A100, H100, B200, MI300X & GPUMode & Organizer review \\
  AlphaEvolve~\citep{novikov2025alphaevolve} & No & Evolutionary program database & TPU/\allowbreak{}GPU & Internal and mathematical & - \\
\end{xltabular}
}

\section{As-Of Dates}
\label{sec:app-j}

{\cktabledense\renewcommand{\arraystretch}{1}
\begin{xltabular}{\linewidth}{@{\hspace{4pt}}W{0.9}W{1.3}W{1.0}W{0.9}W{0.9}@{\hspace{4pt}}}
  \caption{\textbf{Every headline number with its basis and as-of time.} ``Site'' = timed or scored by the official site; ``local'' = our B200 pods (clock unlocked; reads high).}
  \label{tab:asof}\\
  \toprule
  \thead{Number} & \thead{Value} & \thead{Basis} & \thead{As of} & \thead{Note} \\
  \midrule
  \endfirsthead
  \multicolumn{5}{@{}l}{\emph{Table~\thetable{} (continued)}}\\
  \toprule
  \thead{Number} & \thead{Value} & \thead{Basis} & \thead{As of} & \thead{Note} \\
  \midrule
  \endhead
  \bottomrule
  \endlastfoot
  L1 board & \#1, 0.823339 & site, public board & \boarddate{} & collection board; \#1 since 2026-10-27, apart from a short loss on 2026-11-07 \\
  \ckz L2 board & \#1, 0.819260 & site, public board & \boarddate{} & collection board; \#1 from 2026-10-29 to 2026-11-07; \#2 from 2026-11-07 to 2026-11-10 \\
  Quantization board & \quantrank{}, 0.909670 & site, public board & \boarddate{} & collection board \\
  \ckz FlashInfer-Bench & verified correct on the site as private rows (26 of 26); results not published & site, private & kernels as of 2026-11-10 & no score, rank or speedup reported \\
  Per-problem \#1s & 89 of 209 (L1 35, L2 39, Quant. 15) & site, public & \boarddate{} & per-problem boards \\
  \ckz Per-problem top 10 & 193 of 209 (L1 84, L2 76, Quant. 33) & site, public & \boarddate{} & per-problem boards \\
  Site-correct kernels & 235 of 235 & site & kernels as of 2026-11-10 &  \\
  \ckz Speedup vs official baseline & $4.40\times$ geomean (median $3.78\times$), 209 public-board problems & site, our public rows & \boarddate{} & L1, L2 and Quantization only \\
  Per board & L1 $4.04\times$, L2 $3.62\times$, Quant. $9.19\times$ & site, our public rows & \boarddate{} &  \\
  \ckz Speedup vs \texttt{torch.compile} & $11.2\times$ geomean, all 235 & local & kernels as of 2026-11-10 & default mode; a familiar reference, not the benchmark's baseline \\
  Pod-timed speedup vs official baseline & $5.30\times$ & local & kernels as of 2026-11-10 & explains the local--site gap only \\
  \ckz Concurrent agent sessions & over a thousand at peak (controllers plus children) & our traces & final run &  \\
  Kernels tested in the final run & 98.4\% correct & local & 2026-11-02 &  \\
  \ckz Throughput after one warm restart & 3.6$\times$ evaluated candidates per hour & local & matched hours before and after the restart & several changes shipped together; split not measured \\
  Lock time per landed candidate & $-71\%$ & local & the same matched hours &  \\
  \ckz Gain timing & each board gained 73--85\% of its rise in mean best SOL in the first 15\% of the run's time & local & end of the data window & \S\ref{sec:climb} \\
  Robustness coverage & 94.9\% of new bests checked & local & \freezedate{} &  \\
  \ckz Controller vs tree & 35 of 38; 2.42$\times$ correct attempts & local & Phase 1 (pre-final harness) & about even at an equal candidate count (better on 15, worse on 23) \\
  Mandatory references & ratio 0.981 over 233 pairs; 126 vs 107 & local & Phase 2 (pre-final harness) &  \\
  \ckz Quant/030 best path & $6.25\times$ $\to$ $27.36\times$ & local & case-study cut & a later best reached $27.57\times$ \\
  Languages of the final kernels & 214 CUDA C++, 13 CuTe DSL, 6 Triton, 2 mixed & - & kernels as of 2026-11-10 &  \\
\end{xltabular}
}

\FloatBarrier
\section{Dashboard Views}
\label{sec:app-views}

\begin{figure}[!htbp]
  \centering
  \includegraphics[width=\linewidth]{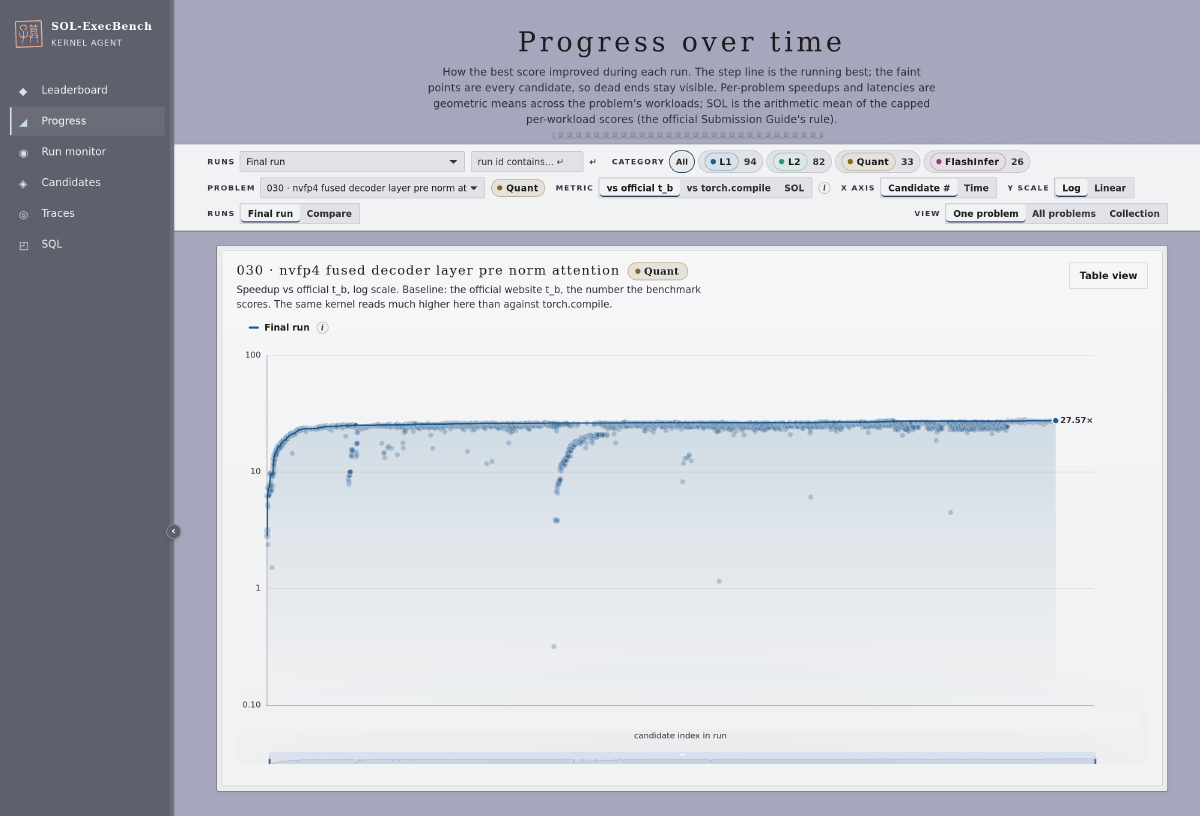}
  \caption{Progress view (a snapshot, so its numbers can differ slightly from the text): how one problem's best grew in the final run. The step line is the running
    best; faint dots are every candidate. The metric can be speedup or SOL, and the x axis the
    candidate number or time. On the SOL view, dashed lines show the public board's top three. Local
    values read higher than the site's score.}
  \label{fig:dash-progress}
\end{figure}

\begin{figure}[!htbp]
  \centering
  \includegraphics[width=\linewidth]{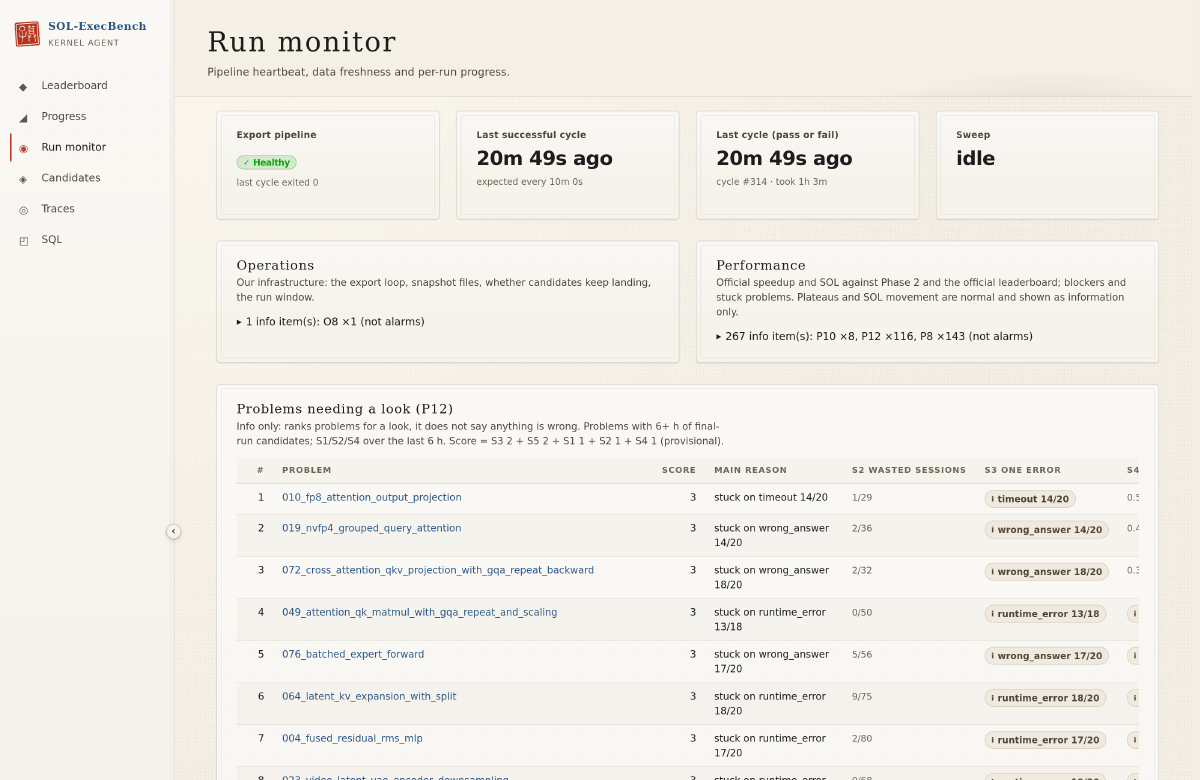}
  \caption{Run monitor view: whether the data export is healthy and how old the data is, the
    monitor's issues for operations and performance by level (decision, warn, info), and the
    problems that may need a look, ranked by a score with the main reason (for example, many recent
    tries stuck on one error). The chat alerts read the same issues. The final run's summary card is
    left out of this shot.}
  \label{fig:dash-monitor}
\end{figure}

\begin{figure}[!htbp]
  \centering
  \includegraphics[width=\linewidth]{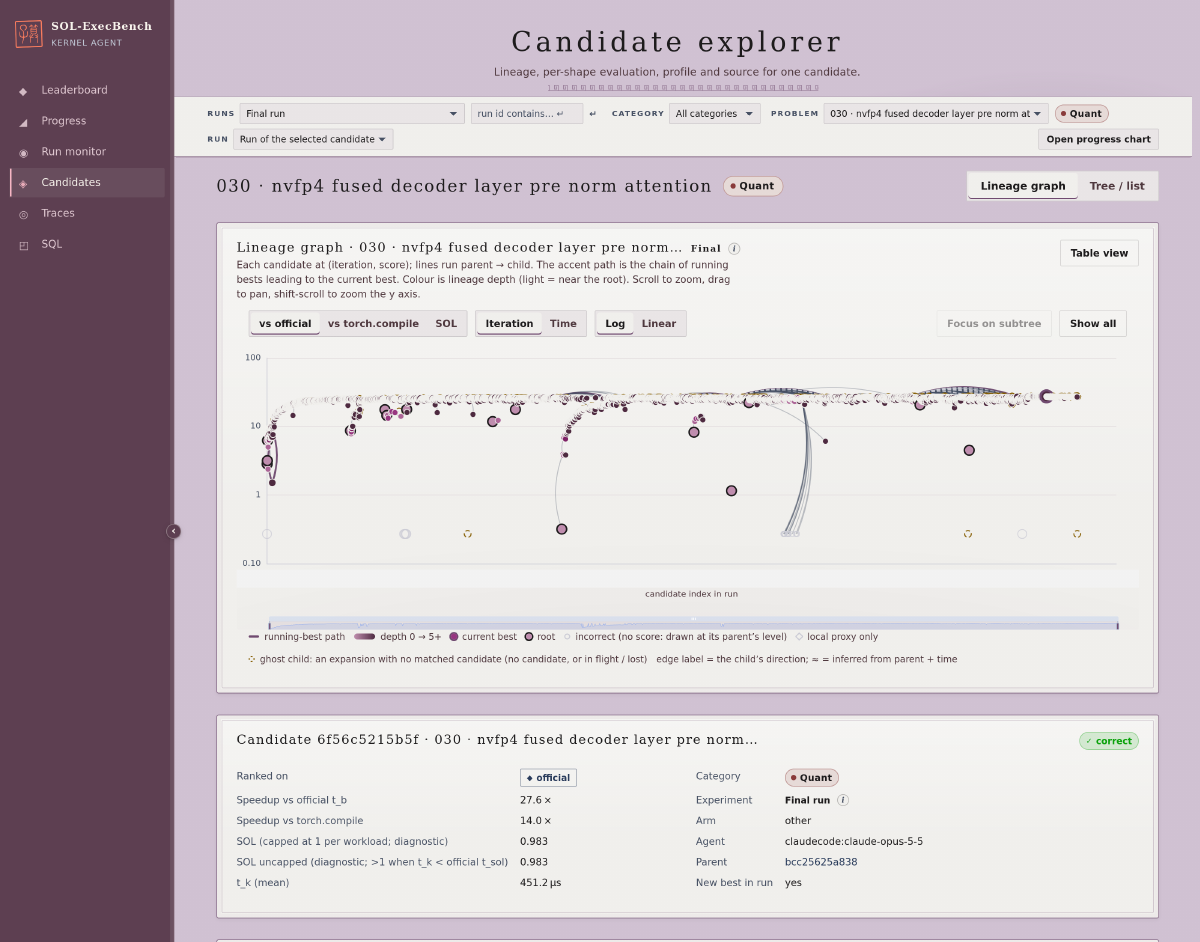}
  \caption{Candidates view: the lineage graph of one problem's run. Each dot is a candidate, placed by
    its order in the run (x) and its speedup over the benchmark baseline (y, log scale). Lines run
    from parent to child; ringed dots are roots, and hollow dots are incorrect rows, drawn at their
    parent's level. A click opens a candidate's speedups, score on each shape, profile, note and
    source.}
  \label{fig:dash-candidates}
\end{figure}

\begin{figure}[!htbp]
  \centering
  \includegraphics[width=\linewidth]{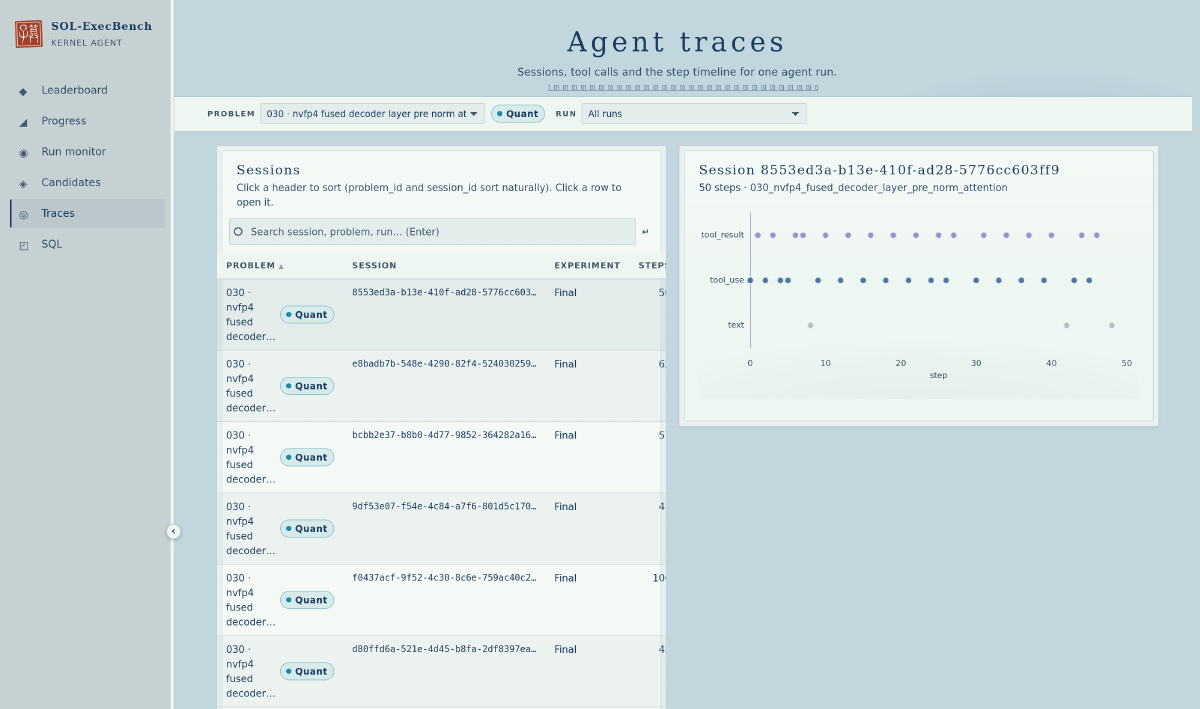}
  \caption{Traces view: the agent sessions of one problem and one session's step timeline (tool
    calls, tool results and text, by step). The view's tool-call chart is left out of this shot;
    Figure~\ref{fig:tool-share} gives each tool's share.}
  \label{fig:dash-traces}
\end{figure}

\begin{figure}[!htbp]
  \centering
  \includegraphics[width=\linewidth]{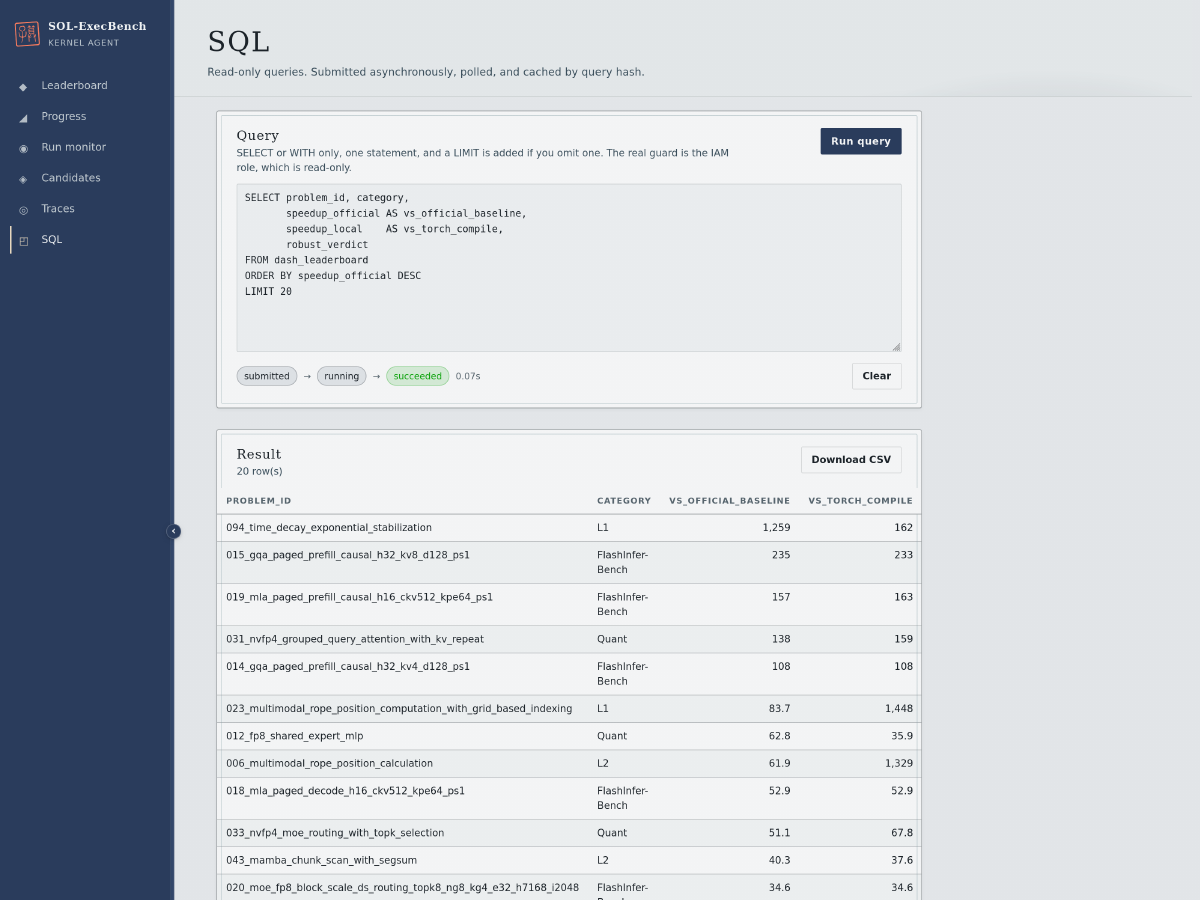}
  \caption{SQL view: one read-only query at a time over the dashboard's data snapshot, and a result
    table. The query shown lists each problem's best kernel with
    its speedup over the benchmark baseline and over torch.compile, and its robustness verdict, both timed on our pods (so they differ
    from the site-timed numbers in \S\ref{sec:boards}).}
  \label{fig:dash-sql}
\end{figure}

The dashboard views of Appendix~\ref{sec:app-h}, as screenshots of the dashboard's own pages
(Figures~\ref{fig:dash-progress}--\ref{fig:dash-sql}). Run ids, paths, host names, internal names and absolute counts are hidden. Each view
is shown in one of the dashboard's color themes. The Leaderboard view is not shown. Each view answers one question.
The \textbf{Leaderboard} view asks what our best valid kernel per problem is, on which baseline,
whether it is robust, and how far it is from the board's \#1; it also shows a calibrated estimate of
the site's score, for people only (\S\ref{sec:reward}). The \textbf{Progress} view shows how
one problem's best grew, by candidate or over time, as speedup or SOL; dashed lines show the board's
top three. The \textbf{Run monitor} shows whether the data export is alive, how old the data is, and
which runs or problems need a look; its issues panel lists the monitor's findings. The
\textbf{Candidates} view shows how the search reached the best: it draws one run's lineage graph, and
a click opens a candidate's speedups, its score on each shape, its profile, its note and its source.
The \textbf{Traces} view shows what the agents spent their turns on: tool calls, sessions and one
session's step-by-step timeline. The \textbf{SQL} view answers any quick read-only question, over
the same data snapshot.

\FloatBarrier
\section{The Public Leaderboards}
\label{sec:app-boards}

\begin{figure}[!htbp]
  \centering
  \includegraphics[width=0.9\linewidth]{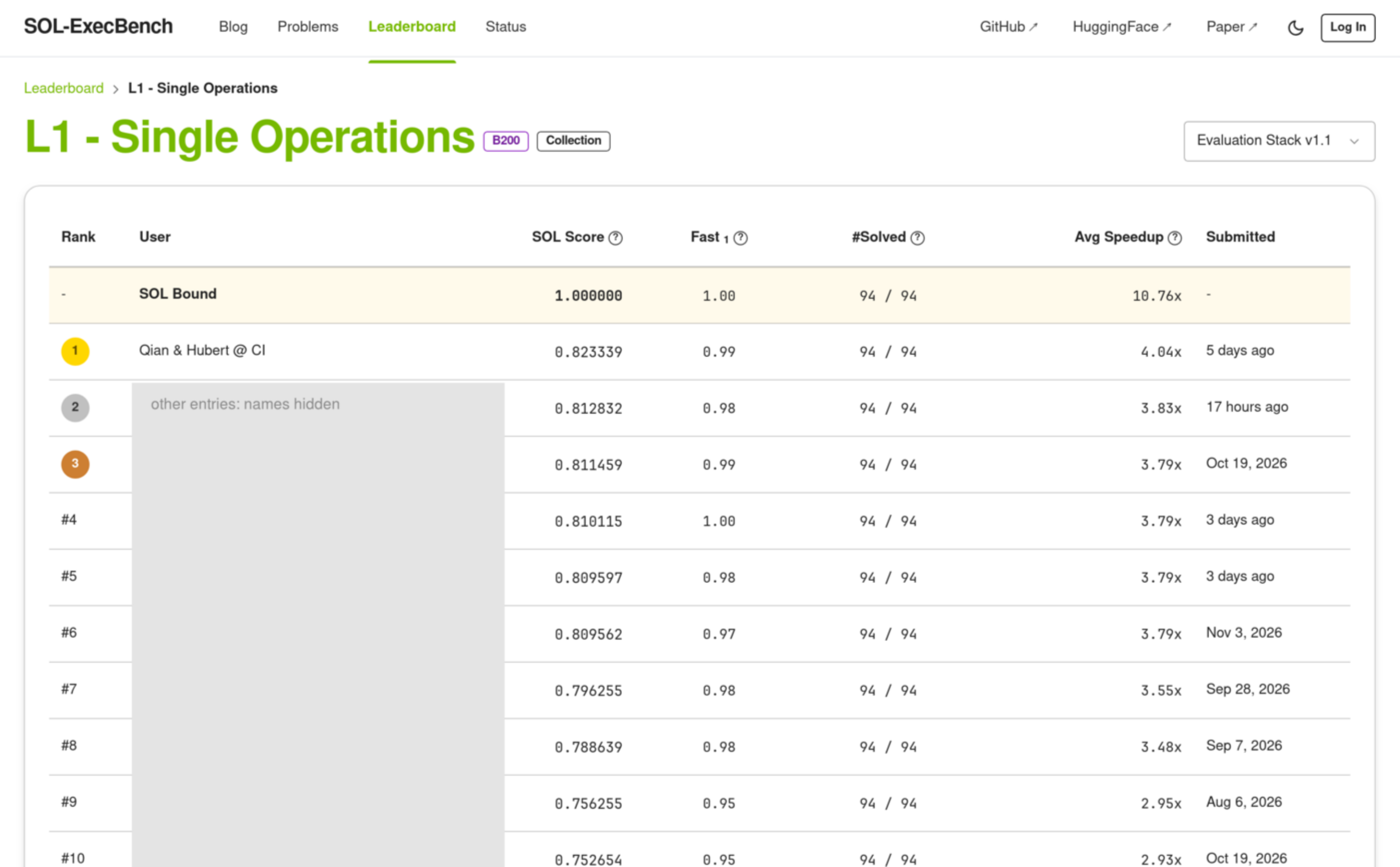}
  \caption{The official SOL-ExecBench L1 leaderboard on B200, captured on 2026-11-11 00:27 UTC; we rank \#1. Archived copy: \url{https://web.archive.org/web/20261111001510/https://research.nvidia.com/benchmarks/sol-execbench/leaderboard/collection/1/B200}.}
  \label{fig:board-l1}
\end{figure}

\begin{figure}[!htbp]
  \centering
  \includegraphics[width=0.9\linewidth]{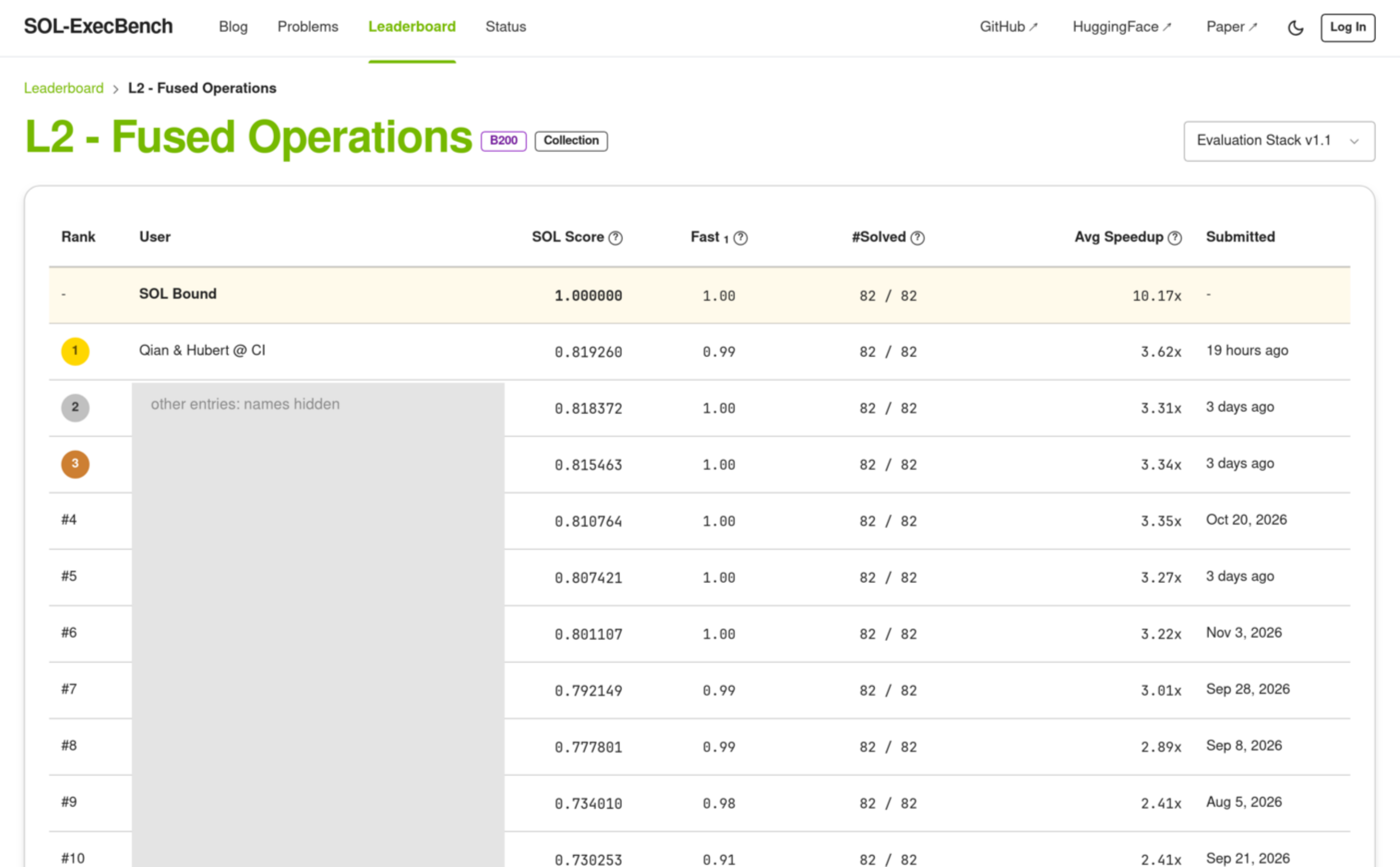}
  \caption{The official SOL-ExecBench L2 leaderboard on B200, captured on 2026-11-11 00:27 UTC; we rank \#1. Archived copy: \url{https://web.archive.org/web/20261111001524/https://research.nvidia.com/benchmarks/sol-execbench/leaderboard/collection/2/B200}.}
  \label{fig:board-l2}
\end{figure}

\begin{figure}[!htbp]
  \centering
  \includegraphics[width=0.9\linewidth]{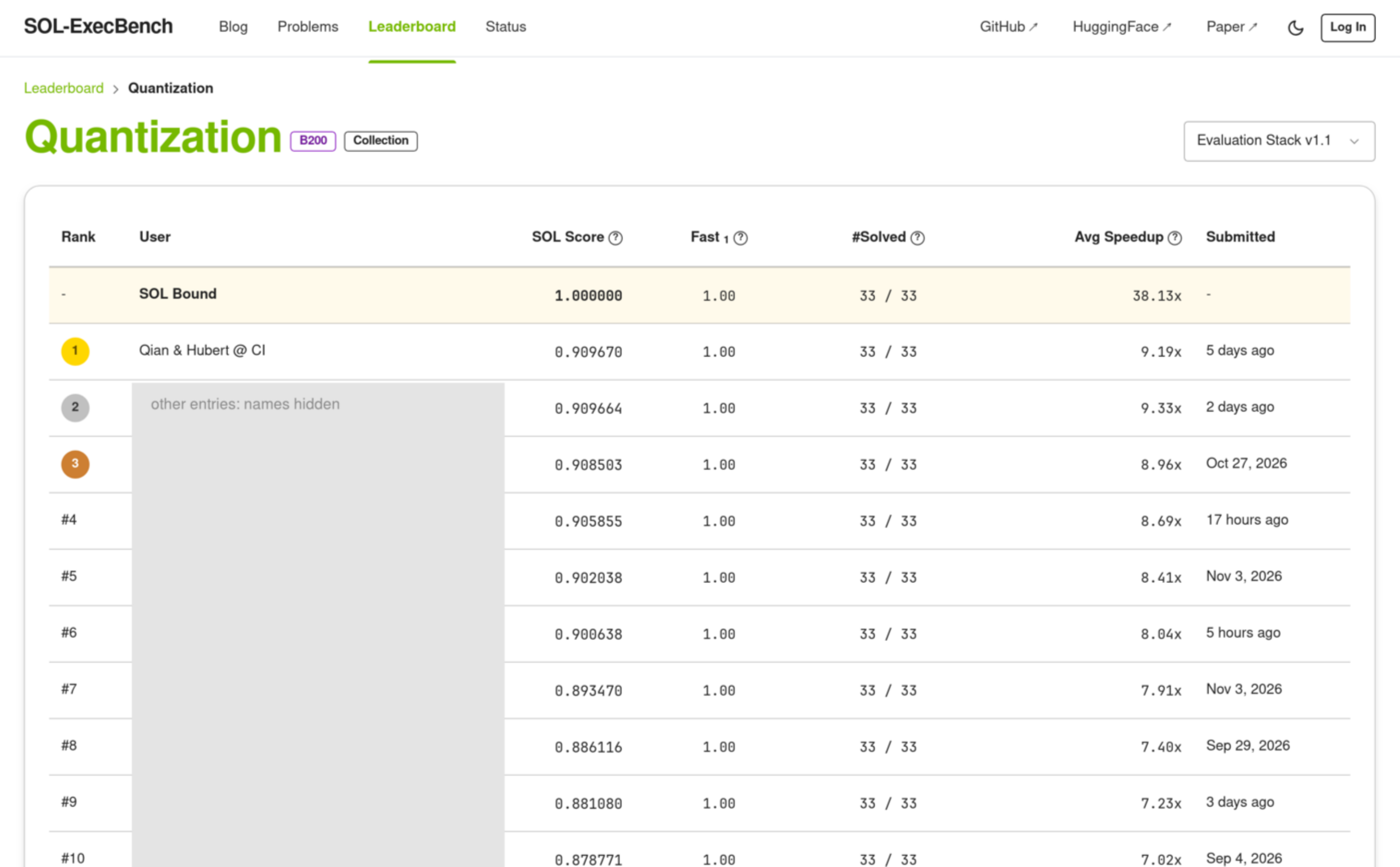}
  \caption{The official SOL-ExecBench Quantization leaderboard on B200, captured on 2026-11-11 00:27 UTC; we rank \#1. Archived copy: \url{https://web.archive.org/web/20261111001536/https://research.nvidia.com/benchmarks/sol-execbench/leaderboard/collection/3/B200}.}
  \label{fig:board-quant}
\end{figure}

Figures~\ref{fig:board-l1}, \ref{fig:board-l2} and \ref{fig:board-quant} show the official
SOL-ExecBench leaderboards for the three boards we submitted to, as captured on 2026-11-11 00:27 UTC. Copies are
kept in the Internet Archive (links in the captions). Figure~\ref{fig:F1} and Table~\ref{tab:boards} give the ranks as of \boarddate{}.

\end{document}